\documentclass{article}

\usepackage[main, final]{neurips_2026}

\usepackage[utf8]{inputenc}
\usepackage[T1]{fontenc}
\usepackage{hyperref}
\usepackage{url}
\usepackage{booktabs}
\usepackage{amsmath}
\usepackage{amssymb}
\usepackage{amsfonts}
\usepackage{nicefrac}
\usepackage{microtype}
\usepackage{xcolor}
\definecolor{aidraft}{RGB}{60,80,120}
\usepackage{graphicx}
\usepackage{subcaption}

\newcommand{\synth}{\textsc{Synth}}

\title{It's All Training: A Fully Synthetic Single-Stage Recipe for LLMs}

\newcommand{\takeaway}[1]{\vspace{1mm}{\color[HTML]{005D96}\textbf{{$\triangleright$\hspace{5pt}#1}}}}

\author{%
\begin{minipage}{0.85\textwidth}\centering
\textbf{Pierre-Carl Langlais*}$^{1,2,3}$ \quad
\textbf{Pieter Delobelle*}$^{1,9}$ \quad
\textbf{Yannick Detrois}$^{1,4}$ \\[2pt]
\textbf{Pavel Chizhov}$^{1,5}$ \quad
\textbf{Carlos Rosas-Hinostroza}$^{1,7}$ \quad
\textbf{Neil Si Smail}$^{1}$ \\[2pt]
\textbf{Benjamin Burtin}$^{1}$ \quad
\textbf{Hanna Shcharbakova}$^{8}$ \\[2pt]
\textbf{Ivan Yamshchikov}$^{1,5}$ \quad
\textbf{Anastasia Stasenko}$^{1,6}$ \\[8pt]
{\normalfont\small
$^{1}$PleIAs\\[2pt]
$^{2}$Sorbonne Center for Artificial Intelligence \quad
$^{3}$Sciences Po Médialab \quad
$^{4}$EPFL \\[2pt]
$^{5}$CAIRO, Technical University of Applied Sciences Würzburg-Schweinfurt \\ [2pt]
$^{6}$Paris Dauphine-PSL \quad
$^{7}$Lattice, ENS-PSL \\[2pt]
$^{8}$TU Munich, Munich Center for Machine Learning\quad
$^{9}$KU Leuven \\[4pt]
Correspondence: \texttt{\{pierre-carl, pieter\}@pleias.ai}}
\end{minipage}%
}

\begin{document}

\maketitle

\begin{abstract}
Current pre-training datasets are derived from web crawls, with all their issues, and were not designed to support mid- and post-training pipelines — for instance, they contain little explicit reasoning. Thus, many frontier labs have begun to develop their own internal datasets, starting from state-of-the-art models, to augment their pre-training data mix, e.g., with reasoning traces to address cold-start problems. While demonstratively effective, none of these datasets are public, and the effect of this so-called \emph{synthetic data} on knowledge and skill acquisition of language models, including small ones, remains poorly understood. We present \synth, the first open-source synthetic corpus derived from 58,698 Wikipedia articles that collapses pre-, mid-, and post-training into a single training stage via structured amplification of curated encyclopedic seeds. We evaluate {\sc Synth} by training a suite of models: a 56M tiny model ({\sc Monad}), 0.3B--0.6B dense models ({\sc Baguettotron}), and a 13B-total / 1B-active Mixture-of-Experts. At iso-compute, {\sc Synth} outperforms filtered web data, and our models remain competitive with similarly-sized open-weight baselines. Because {\sc Synth} is back-translated from grounded passages, {\sc Synth}-trained models achieve high factual precision despite 10--140$\times$ fewer training tokens, with memorization targeted by the seed corpus. These results show that synthetic datasets, including our {\sc Synth} dataset, are capable of producing competitive generalist models from a fraction of the training data, enabling rapid iteration as the frontier advances. These findings open up possibilities for both generalist models with significantly increased data efficiency, as well as domain-specific models where no instruction  or conversational data is available. Finally, we publicly release our {\sc Synth} dataset and the suite of {\sc Baguettotron} models under a permissive license, thus supporting open-source language model development.
\end{abstract}
\begin{center}
\raisebox{-0.2em}{\includegraphics[height=1.2em]{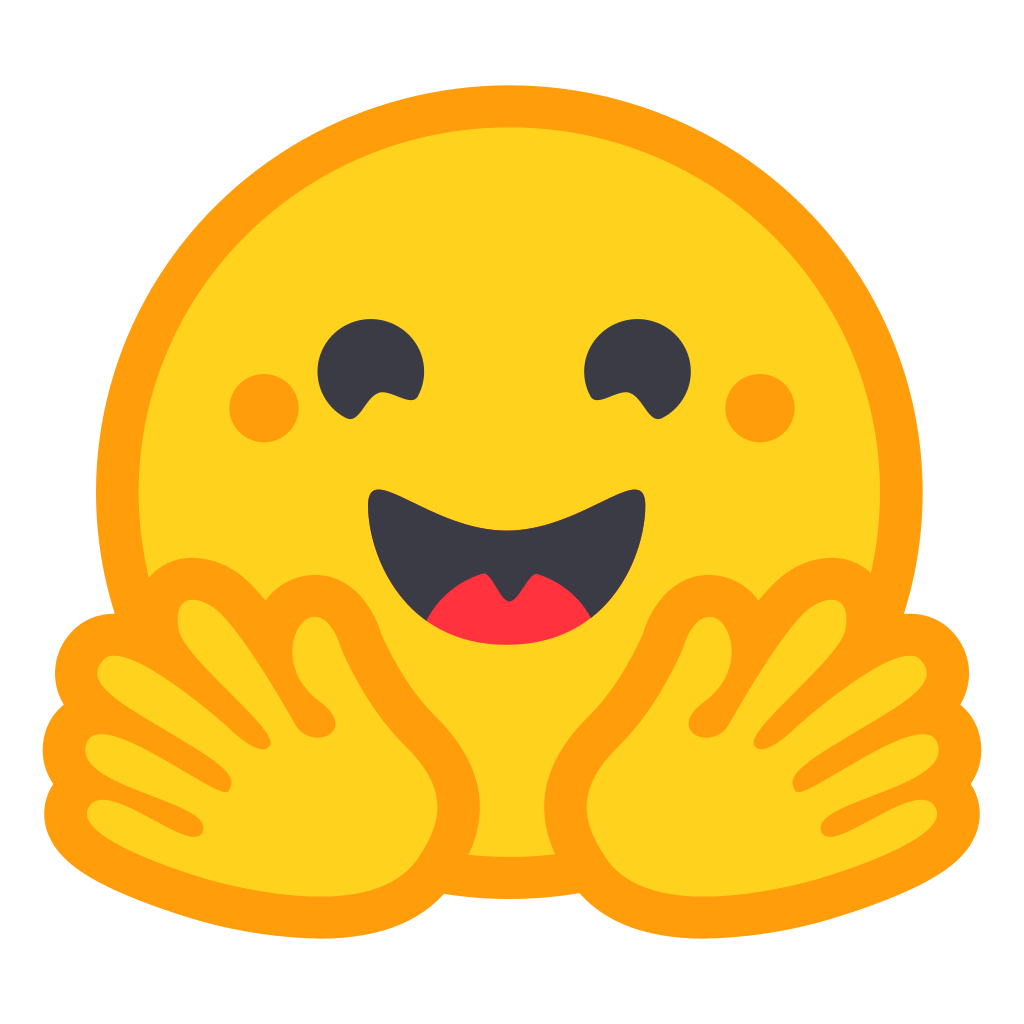}}~\href{https://huggingface.co/datasets/pleias/Synth}{\texttt{pleias/synth}} \quad | \quad \raisebox{-0.2em}{\includegraphics[height=1.2em]{hf-logo.png}}~\href{https://huggingface.co/pleias/Baguettotron}{\texttt{pleias/baguettotron}}
\end{center}

\section{Introduction}
Crawled data has been the most prominent data source for Large Language Model (LLM) pre-training, for instance, for the Pile~\citep{pile} or Common Crawl. 
However, this data allows for little control over content.
Furthermore, in response to \emph{AI scraping}, some site owners have taken countermeasures to prevent their content from being included in pre-training datasets.
As such, most labs have diversified their training mix, for instance, including digitized cultural heritage and scientific literature~\citep{langlais2026common} or synthetic data, which is the focus of this work.

Prior work on synthetic data has explored rephrasing \citep{maini2024rephrasing, maini2025beyondweb} and quality filtering \citep{idahl2026propella, li2024datacomp, su_nemotron-cc_2025} of existing corpora.
Labs are also incorporating reasoning traces and other high-quality LLM outputs into pre-training mixes, e.g.\ from DeepSeek's R1 model.
However, this amounts to distillation with little control over the input beyond the initial query.
This approach also heavily depends on source-model quality: unverified errors can enter the mix, and recursive training risks \emph{model collapse}, reducing distributional support and output diversity~\citep{shumailov2024curserecursiontraininggenerated,dohmatob2024a,alemohammad2023self}.

We present \synth, a method that addresses these issues by \emph{back-translating} instruction-tuning data (with RAG support and reasoning) from a set of controlled \emph{seeds}, in our case 58k multilingual Wikipedia articles.
This way, we control which facts the models memorize and the language distribution of the training corpus. By including back-translated reasoning and instructions, we can also eliminate separate training stages for instruction tuning or reasoning entirely: the models presented in this paper undergo \textbf{no} supervised fine-tuning or reinforcement learning.
Seed grounding and task heterogeneity also mitigate model collapse by construction~\citep{gerstgrasser2024model, fengbeyond}. To summarize, we make the following contributions in this work:
\begin{itemize}
    \item We present an open-source dataset, \synth, comprising almost 80B tokens of synthetic text  in 8 languages, seeded from 58k Wikipedia articles, suitable for all training stages (\autoref{ch:synth}).
    \item We train a suite of models on \synth, ranging from 0.05B dense to 13B Mixture-of-Experts, achieving competitive performance with similarly sized open models including Gemma, Qwen, and LFM across a wide range of tasks (\autoref{ch:baguettotron}).
    \item We extensively investigate factual recall from seed articles, showing that our suite of models outperforms comparable open models at memorizing target knowledge (\autoref{ch:memorization}).
    \item We show that our method also applies to domain or task adaptation and continuous learning settings by creating a telecommunications-specific model with significant improvement over domain-specific evaluations (\autoref{ss:tele}).
\end{itemize}

\section{Background and related work}\label{ch:related}

Since GPT-3, public LLM research has primarily relied on web archives, for instance, C4~\citep{raffel2020exploring}, ROOTS~\citep{laurencon2023bigsciencerootscorpus16tb}, and FineWeb~\citep{penedo2024the,penedo2025fineweb2pipelinescale}. Dolma~\citep{soldaini2024dolma} and The Pile~\citep{pile} expand beyond web crawls to curated educational materials, books, and other domains. However, data reproducibility has been impeded by structural liability concerns, as most crawled web pages are covered by copyright. This leads to content removals~\citep{kandpal2025the,cooper_extracting_2025}, licensing confusions~\citep{longpre_data_2023} and general weakening of data commons~\citep{longpre_consent_2024}. Recent pre-training datasets enforce stronger openness and provenance, including KL3M~\citep{bommarito_kl3m_2025}, Common Pile~\citep{kandpal2025the}, and Common Corpus~\citep{langlais2026common}. The hybridization of open data sources and instruction datasets has been shown to effectively bridge the performance gap for small models, thereby demonstrating the viability of a fully open approach~\citep{nguyen_mixturevitae_2026}.

\textbf{Synthetic (pre-)training.~~}
Advances in reasoning and agentic models have spurred the use of synthetic text across all training phases. \textit{Mid-training} employs more compute-intensive methods than standard post-training, scaling structured instructions, code simulations, and agent traces to billions of tokens~\citep{abdin_phi-4_2024,walsh2025,mo_mid-training_2025}. Synthetic generation helps engineer specific capabilities and knowledge bases that are scarce in publicly available datasets~\citep{liu_best_2024,davidson_orchestrating_2025,mullahmetov_synthetic-based_2025}, especially for long context~\citep{kim_data-efficient_2026}. \textit{Open-Thoughts} provides a rare public mid-training dataset, featuring 1.2M samples with extensively documented ablations~\citep{guha_openthoughts_2025}.

\textit{Synthetic rephrasing} reshapes real data sources (\textit{seeds}) at scale. This enables selective amplification, memorization, and stylistic/semantic improvement. \citet{maini2024rephrasing} show that quality-targeted rewriting of web seeds supports memorization better than literal repetition, a finding extended by BeyondWeb~\citep{datologyai_beyondweb_2025} and by the high-quality synthetic subset of Nemotron-CC~\citep{su_nemotron-cc_2025}. BeyondWeb also reports diminishing returns past $\sim$3B-parameter rephrasers under fixed compute, motivating small specialized generators over a single large one. Many recent major model releases rely on some form of synthetic rephrasing~\citep{minimax_minimax-m1_2025,singh_arcee_2026,kimiteam2026kimik2openagentic,kimiteam2026kimik25visualagentic,nvidia_nvidia_2025}.

Fully-synthetic training (or \textit{synthetic pre-training}) is a more restricted category. Phi-1.5~\citep{li_textbooks_2023} trained a 1.3B parameter model on a corpus dominated by synthetic textbooks, matching the scores of much larger models; Cosmopedia-1b~\citep{benallal2024cosmopedia} reproduced the recipe in the open. Reliance on a single generation template (textbook rewriting) has left these works vulnerable to the surface-diversity collapse~\citet{kang_demystifying_2025}, and subsequent Phi releases reintroduced curated web data. We revisit synthetic pretraining and replace single-template generation with a constraint grammar over heterogeneous task pipelines explicitly targeting that failure mode.

\textbf{Controlled environments.~~}
Synthetic data facilitates reproducible training experiments by enabling clean, controlled experiments in playgrounds and engineered corpora that avoid real-world noise and contamination~\citep{allen-zhu_physics_2024,AllenZhu-icml2024-tutorial}. Such environments are especially prevalent in research on LLM memorization. \citet{morris_how_2025} estimate the pure memorization capacity of GPT-style transformers at $\sim$3.6 bits per parameter by training from scratch on uniformly random bit-strings, where no generalization is possible; \citet{allen-zhu_physics_2024} report a complementary $\sim$2 bits per parameter on synthetic biographies and phone books, the gap recovered through generalization. This capacity is highly sensitive to data composition: a 1:7 useful-to-junk ratio reduces effective capacity for useful knowledge by a factor of 20~\citep{allen-zhu_physics_2024}, providing direct theoretical motivation for the seed-curated approach we adopt. \citet{ye_cram_2026} show that a 110M model pre-trained on annotated Wikipedia matches a $10\times$ larger baseline on entity-fact recall.
At a larger scale, \citet{lin_learning_2025} show that Llama 3.1 8B, mid-trained on 1T tokens of synthetic Wikipedia-sourced study material, beats much larger baselines on closed-book factual QA. 

\section{\textsc{Synth}}\label{ch:synth}

\begin{figure}[t]
    \centering
    \includegraphics[width=\linewidth]{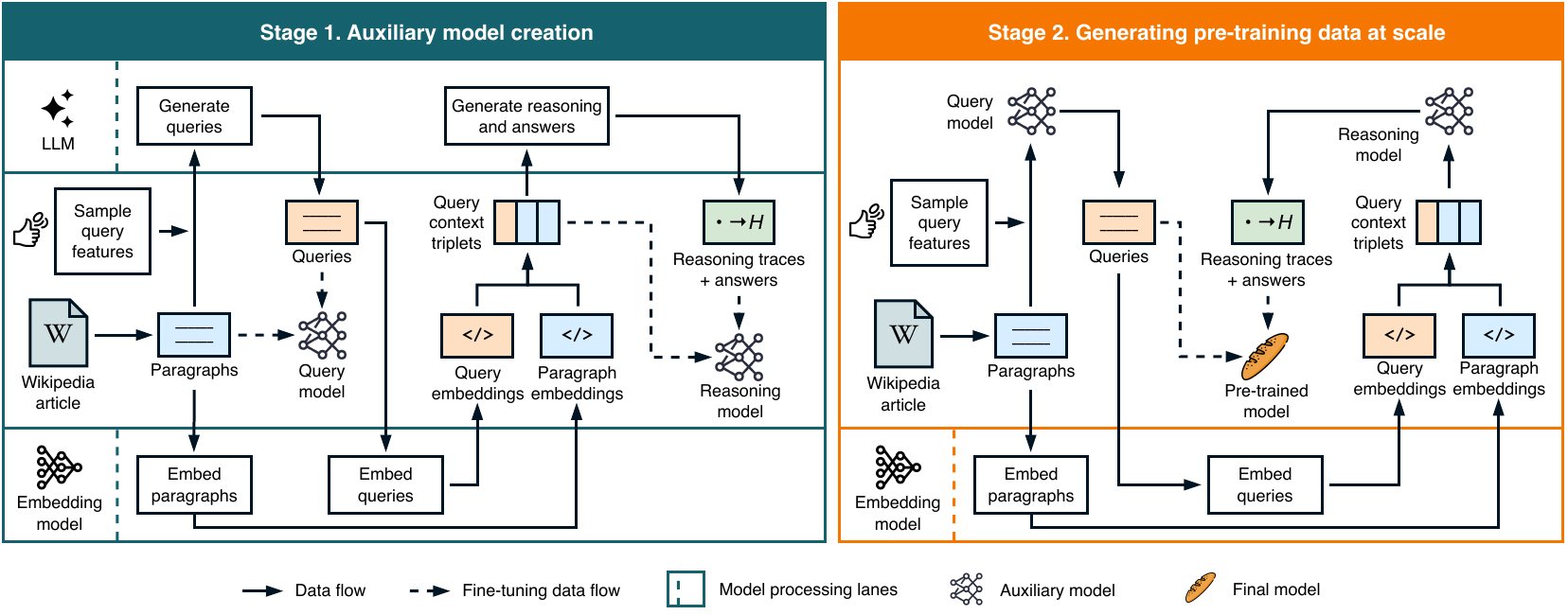}
    \caption{\textbf{\textsc{Synth} pipeline for the memorization task.} At Stage 1, auxiliary models are fine-tuned on LLM-generated data. At Stage 2, these models are used to produce synthetic training data at scale.}
    \label{fig:placeholder}
\end{figure}

{\sc Synth} amplifies a fixed set of curated seed segments through a two-stage pipeline (\autoref{fig:placeholder}). \emph{Stage~1} fine-tunes two auxiliaries --- a query model and a reasoning model --- on supervision distilled from a frontier LLM; a frozen \texttt{bge-m3} encoder handles nearest-neighbor retrieval over seed paragraphs. \emph{Stage~2} runs both auxiliaries at scale: each seed is sampled into queries under a constraint-prior model, paired with a retrieved neighbor paragraph, and routed through a task-specific output adapter (memorization QA, RAG, arithmetic, creative writing, editing). 

\textbf{Seeding corpus.~~}
The seed set is anchored on the 50k Wikipedia vital articles (level-1 to level-5\footnote{\url{https://en.wikipedia.org/wiki/Wikipedia:Vital_articles/Level_5}}), community-curated for broad importance-tiered coverage. We add 8{,}698 specialized articles in law, medicine, and chemistry (category-tree and Wikidata graph expansion to fill the gaps surfaced during intermediary evaluation); 3{,}727 Wikibooks pages, primarily cooking and practical knowledge underrepresented in the encyclopedia; and 130 documents covering model self-documentation, recent events, and AI research postdating the snapshot.

\subsection{Memorization}\label{ss:memorization}

\textbf{Queries.~~} The query model is the entry point of Stage~2; any diversity failure here would propagate to the whole \synth{} corpus. Prompting a single large model with rotating instructions empirically collapses to a narrow distribution over phrasings and formats, so we instead fine-tune a LoRA adapter on Gemma-3-12B-base~\citep{gemma_2025} during Stage~1 on curated $\langle$seed, constraints, query$\rangle$ triplets and run in Stage~2 over every seed segment ($\sim$80M tokens, 58k seed articles). Constraints are sampled per call from a simple probabilistic model with independent priors over six axes (query type, complexity, user profile, query result, target language, query style; full enumeration and priors in \autoref{app:query-constraints}). The priors force rows to differ along controlled axes rather than along whichever axes the base model finds easiest to vary.

Every seed segment is uniformly amplified by $100\times$. Importance weighting is implicit in the section-decomposition step: levels 1--4 (top 10k) are split into all their structured sections, while level 5 (IDs 10k--50k) contributes only the lead abstract as a single seed. 

The \emph{query result} axis is the design choice worth flagging: $20\%$ of rows target a refusal, correction, or hedge rather than a confident answer (negative, absurd, or ambiguous queries). Without this signal, a back-translated corpus teaches the model to always answer, since every training query has a confident grounded target; we return to this in \autoref{ch:memorization}, where the FActScore precision gap partly traces to the model declining to confabulate on entities it does not know.

\textbf{Answer and reasoning traces.~~}
The reasoning model is a separate auxiliary (\autoref{fig:placeholder}) that produces an answer along with a back-translated reasoning trace. Its input is a \emph{query-context triplet}: the query, its seed paragraph, and the closest non-seed neighbor retrieved with off-the-shelf \texttt{bge-m3} + FAISS IVF-flat cosine over the seed pool. Stage~1 fine-tunes on triplets paired with traces and answers distilled from a frontier LLM; Stage~2 runs on every Stage-2 query under the same triplet format. Pairing each query with its seed \emph{and} a near-neighbor diversifies grounding combinations and creates semantic bridges that no single-paragraph grounding would produce. The trace uses a stenographic syntax rather than natural-language CoT, with three marker families --- \emph{logical} ($\rightarrow$, $\circlearrowleft$, $\therefore$), \emph{epistemic} ($\bullet$ certain $\ldots$ $\bigcirc$ uncertain), and \emph{verification} ($\square$ / $\boxdot$ / $\checkmark$) --- plus a tree decomposition for multi-step problems. The dedicated tokens (added to the tokenizer, \autoref{ch:baguettotron}) compress a reasoning step into one or two tokens, letting a small model emit a richer trace within a fixed context window. We also insert simulated entropy markers $\langle H{\approx}X.X\rangle$~\citep{xjdr2024entropix} at decision points; these are training-time annotations only, with no inference-time temperature control.

\begin{figure}
    \begin{subfigure}[b]{0.48\textwidth}
        \includegraphics[width=\textwidth]{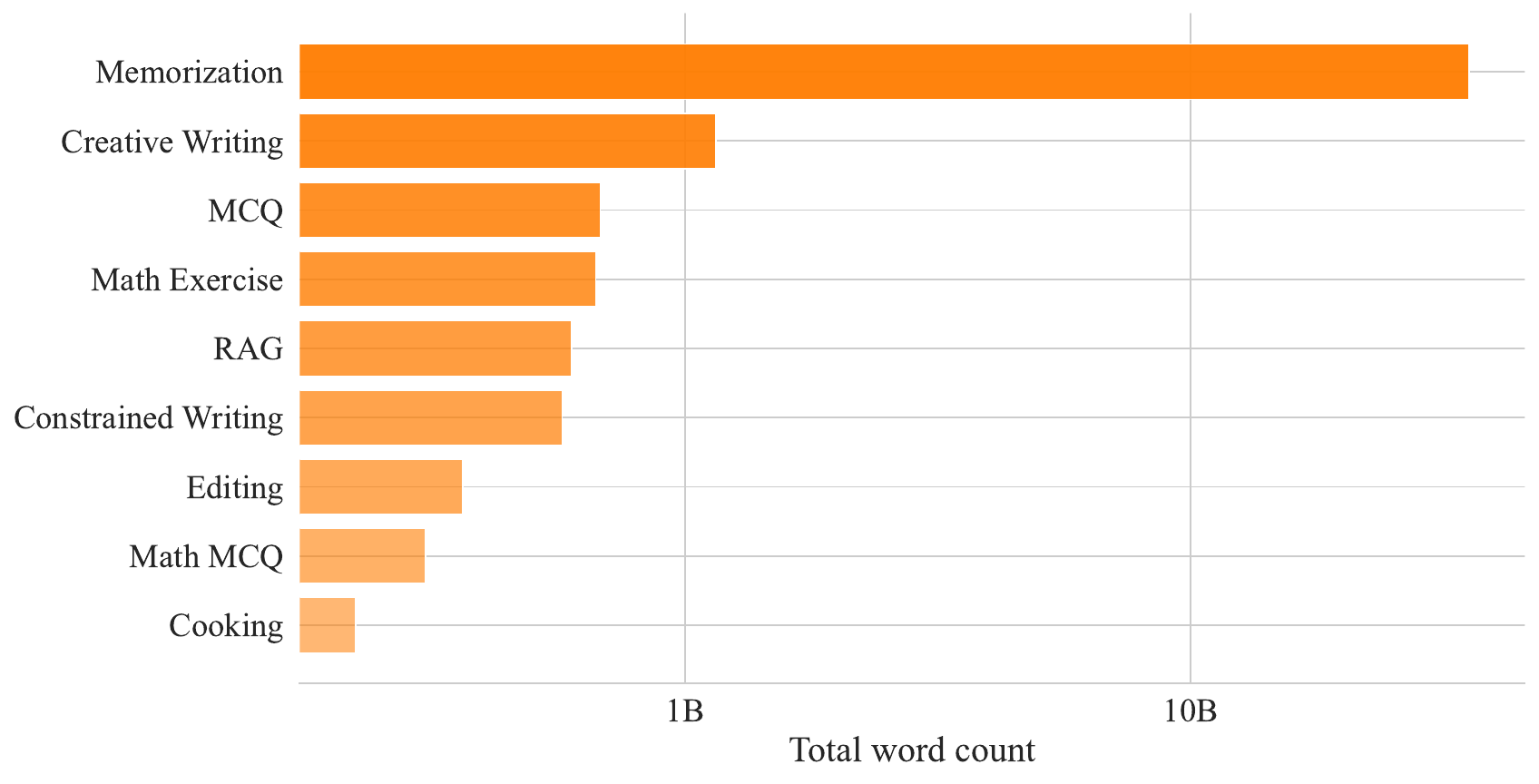}
        \subcaption{Word count distribution for \textsc{Synth} tasks.}
        \label{fig:words-task}
    \end{subfigure}
    \hfill
    \begin{subfigure}[b]{0.46\textwidth}
        \includegraphics[width=\textwidth]{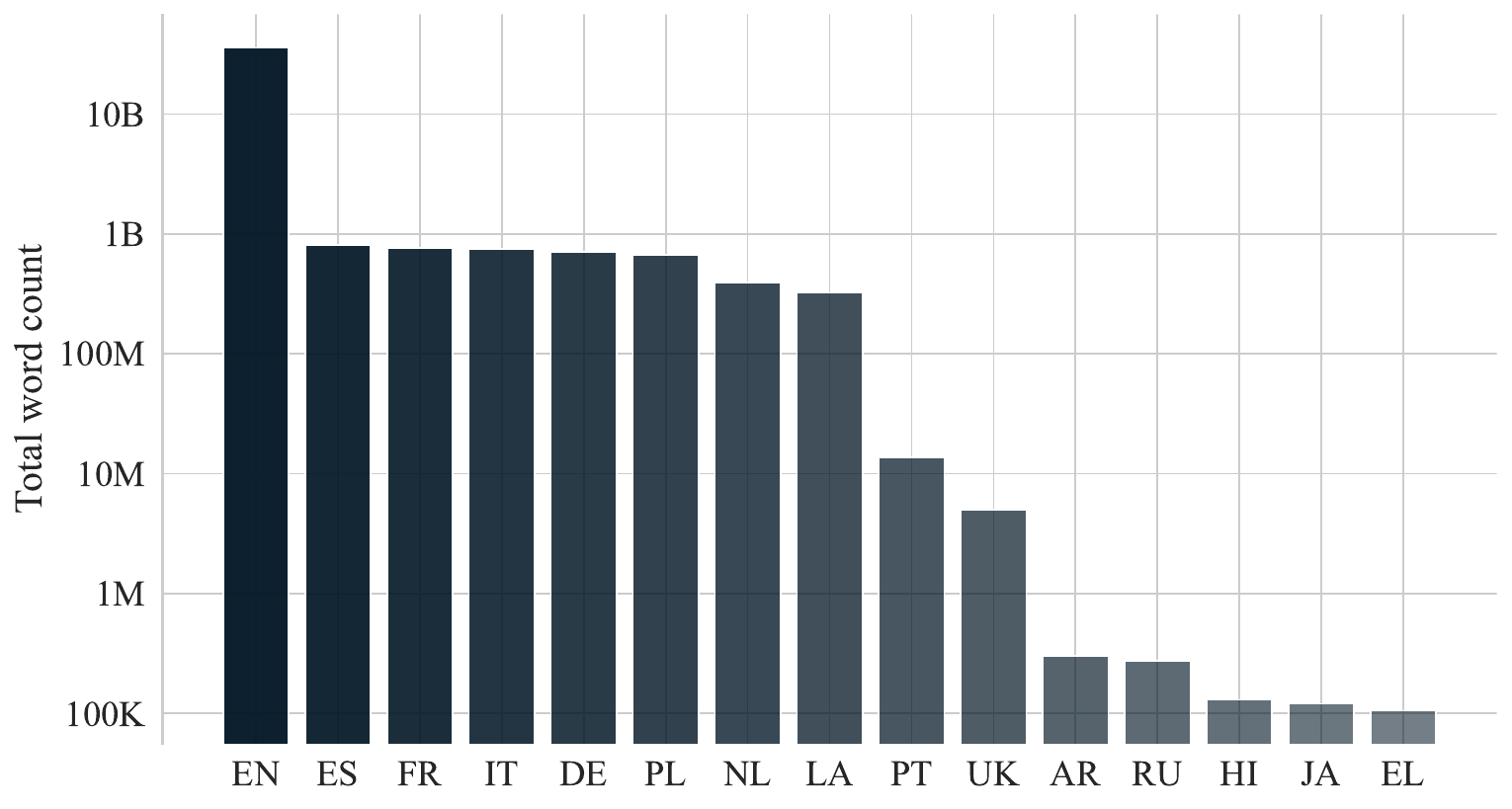}
        \subcaption{Word counts for top-15 \textsc{Synth} languages.}
        \label{fig:words-lang}
    \end{subfigure}
    \caption{Log-scale word counts for \textsc{Synth} tasks and languages.}
    \label{fig:wordcount}
\end{figure}

\subsection{Auxiliary tasks}\label{sec:auxiliary-tasks}
We add task pipelines around memorization, all sharing the constraint-driven flow with task-specific prompts and targets (mix shown in \autoref{fig:words-task}): (1) \emph{RAG} replays the memorization query stream with up to ten retrieved passages and \texttt{<source>}-cited targets, training closed- vs.\ open-book switching; (2) \emph{arithmetic} amplifies ${\sim}3{,}000$ Kimina~\citep{wang2025kimina} templates by randomizing variable values and re-evaluating the symbolic solution (Qwen-3-8B~\citep{qwen3technicalreport}, more accurate than Gemma-3-12b on math), reusing the stenographic syntax of \autoref{ss:memorization}; (3) \emph{creative writing} samples constraint specs (lipograms, layout poems, style/persona) independently of topical seeds; (4) \emph{editing} covers translation, structured extraction, orthographic/factual correction, and style reformulation; (5) \emph{MCQ} adds closed-set questions with distractors; (6) \emph{practical knowledge} runs the main generator over 3{,}727 Wikibooks cooking recipes.

\subsection{Dataset composition}

\textbf{Multilingual generation.~~}
About $20\%$ of \synth{} is non-English, dominated by a handful of European languages selected from Common Corpus (\autoref{fig:words-lang}).
Reasoning traces remain English even for non-English queries, with possible verbatim query quotes.

\begin{figure}
\centering
\includegraphics[width=\linewidth]{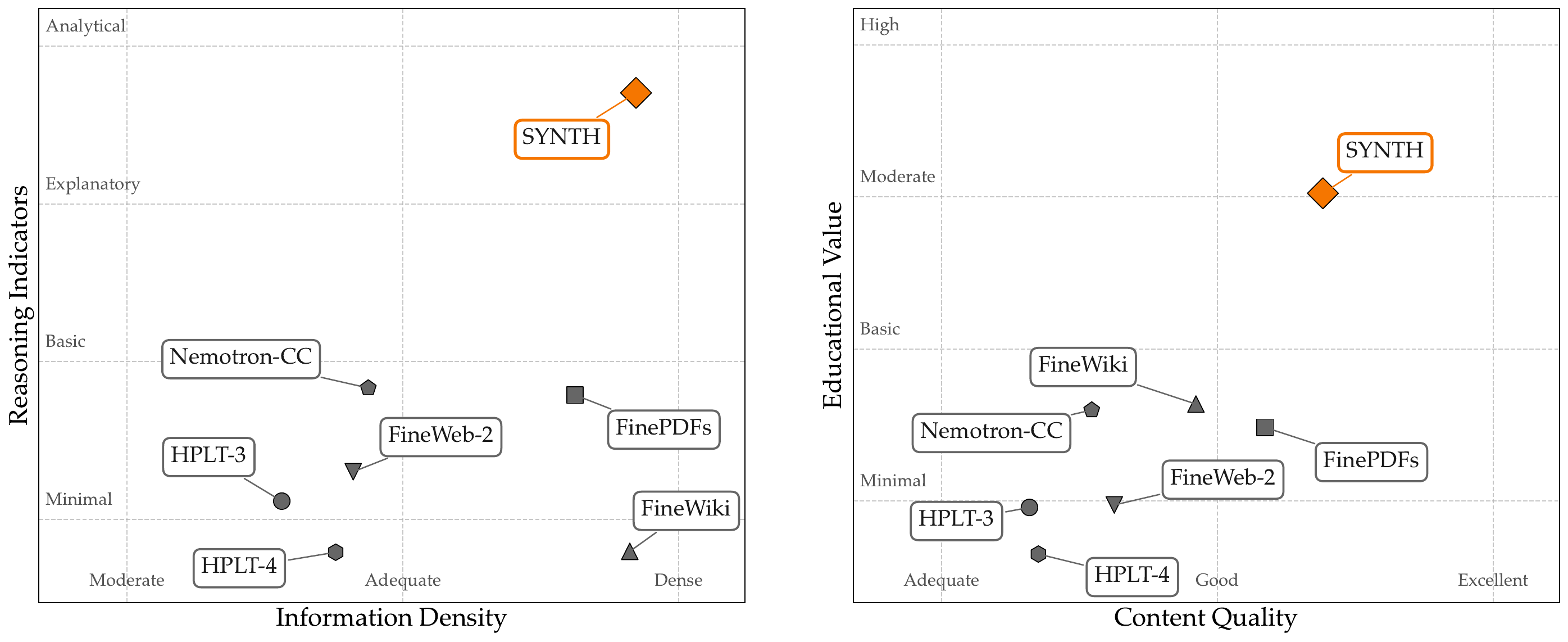}
\caption{\textbf{\synth{} excels in the quality/value pairings by a wide margin} compared to other open pre-training corpora, scored by Propella-1. We present integrity/safety evaluations in Appendix~\ref{app:propella}.}
\label{fig:propella}
\end{figure}

\textbf{Data quality analysis.~~} We use the annotations from Propella-1 multilingual document annotator~\citep{idahl2026propella}, an external data quality judge, to compare \synth{} to a range of open pretraining corpora (including Nemotron-CC~\citep{su_nemotron-cc_2025}, FinePDFs~\citep{kydlicek2025finepdfs}, FineWeb-2~\citep{penedo2025fineweb2pipelinescale}, FineWiki~\citep{penedo2025finewiki}, HPLT 3.0~and 4.0~\citep{oepen-etal-2026-very})\footnote{For FinePDFs, FineWeb2, and HPLT-4, we use language-proportional samples of one million documents each.}. \autoref{fig:propella} reports four property pairings: \synth{} occupies the top-right of both quality / value axes, with the largest margins on \emph{reasoning indicators} and \emph{educational value}. \synth{} also ties FineWiki at the ceiling on integrity / safety axes of the same annotation model (see evaluations in \autoref{app:propella}). The reasoning-indicators score is partly a consequence of the explicit reasoning traces in \synth{}, but the educational-value gap is trace-independent. Wikipedia-derived data matches \synth{} on safety but loses on reasoning; web-cleaned corpora show the opposite tradeoff.

\section{The \textsc{Baguettotron} model suite}\label{ch:baguettotron}

We present three dense models and one Mixture-of-Experts variant.
\begin{itemize}
    \item 
{\sc Monad} (56M parameters, 64 layers, $d_\text{model}=384$) is the smallest configuration and serves as a stress test of how far engineered synthetic data can push a parameter-constrained model.
    \item 
{\sc Baguettotron}-350M (321M parameters, 80 layers, $d_\text{model}=576$) is a deep design, motivated by the hypothesis that deeper stacks could benefit more from reasoning data.
    \item 
{\sc Baguettotron}-600M (594M parameters, 48 layers, $d_\text{model}=1024$) is a more tradition architecture inspired by Qwen and is the main reference run for the rest of the paper.
    \item 
{\sc Baguettotron}-MoE (13.2B total / 1.05B active parameters, 47 layers, $d_\text{model}=1280$, 16 experts top-1, expert hidden $4480$, no shared expert), routes through softmax gating with a $10^{-3}$ load-balance auxiliary loss.
\end{itemize}

\textbf{Tokenizers.~~}
{\sc Baguettotron} uses a BPE tokenizer with a vocabulary size $V=65.535$, trained on a sample from Common Corpus~\citep{langlais2026common} in order to preserve multilingual coverage. The last ${\sim}50$ entries of the vocabulary are {\sc Synth}-specific reasoning control tokens.
{\sc Monad}, the smallest model, instead uses an 8k-vocabulary tokenizer trained on the English section of {\sc Synth}.

\textbf{Infrastructure.~~} Early architecture exploration (depth ablations, {\sc Monad}) used Nanotron on H100 GPUs; other runs use \texttt{torchtitan} with FSDP on 16$\times$H100 GPUs (4 nodes $\times$ 4 GPUs, 64\,GB each). The MoE run uses no expert parallelism; experts are sharded purely through FSDP, as expert parallelism resulted in slightly lower MFU.

\textbf{Training runs.~~} All runs share sequence length $2048$, AdamW (weight decay $0.01$, gradient clipping $1.0$), and a $16.6\%$ linear decay tail to $0.2\%$ of peak lr; per-run details in \autoref{app:training-runs}. {\sc Baguettotron}-MoE matches dense {\sc Baguettotron}-600M's final cross-entropy of $1.19$ on ${\sim}32\%$ as many tokens (${\sim}50$B).

\begin{figure}[t]
\centering
\includegraphics[width=\linewidth]{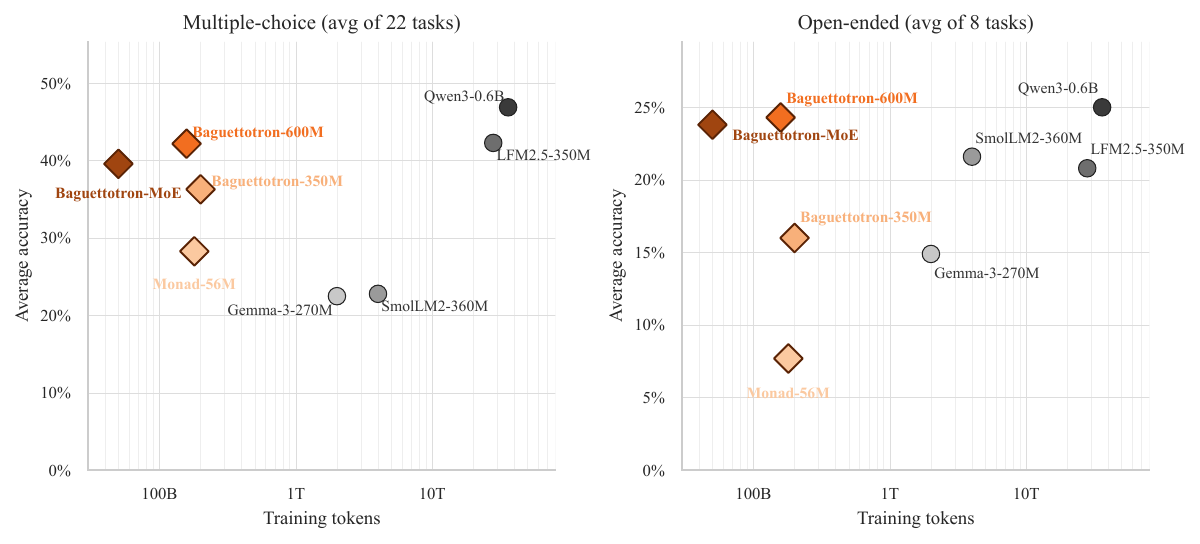}
\caption{\textbf{\textsc{Baguettotron} sits on the token-efficiency frontier, matching or trailing baselines trained on 80--700$\times$ more tokens.} Average accuracy vs.\ pre-training tokens (log scale) on 22 multiple-choice (left) and 8 open-ended (right) tasks. Diamonds: {\sc Baguettotron} (ours); circles: open baselines. Token budgets are publicly reported pre-training token counts (post-training tokens excluded). Top-left is better, i.e. more accurate per token of training data.}
\label{fig:benchmarks-pareto}
\end{figure}

\textbf{Token budget and scaling laws.~~} The Chinchilla-optimal token budget for a 600M model is ${\sim}12$B tokens, $20\times$ parameter count~\citep{hoffman2022scaling}; we trained on $264\times$, deliberately overtrained by web-data conventions. Training loss was still decreasing at step 151{,}000, with no divergence and no loss spikes.
While preliminary, comparing scaling-law fits across three iso-compute $600$M runs of {\sc Synth} against scraped baselines FineWiki and FinePDFs-Edu shows a lower asymptotic loss floor on synthetic data. As loss is not directly comparable across corpora (FineWiki reaches its low loss through ${\sim}17$ epochs over ${\sim}9$B tokens), \autoref{ss:controlled}  compares the same runs on downstream tasks.

\subsection{Evaluation}
\autoref{fig:benchmarks-pareto} compares {\sc Baguettotron} against four open 270M--600M models (Gemma-3-270M~\citep{gemma_2025}, SmolLM2-360M~\citep{allal2025smollm2}, LFM2.5-350M~\citep{amini2025lfm2technicalreport}, Qwen3-0.6B~\citep{qwen3technicalreport}) on 22 multiple-choice and 8 open-ended tasks. Per-benchmark scores, grouped capabilities, and the setup are in \autoref{app:benchmarks}. These baselines differ from ours in architecture, tokenizer, data, and post-training, so we report them as context. 

{\sc Baguettotron}-MoE and {\sc Baguettotron}-600M sit on the token-efficiency frontier on both task types: ours trail Qwen3-0.6B by 4.7 points on multiple-choice and just 0.7 points on open-ended, while training on 80--700$\times$ fewer tokens. {\sc Monad}-56M, despite seeing only 180B tokens at 56M parameters, also sits on the multiple-choice frontier: outperforming Gemma-3-270M and SmolLM2-360M on average. Per-benchmark variance is high: {\sc Baguettotron}-600M reaches 45\% on NuclearQA (vs.\ Qwen3-0.6B's 53\%) and our models tie or beat Qwen on TruthfulQA, ESGenius, and FormationEval, but lag substantially on benchmarks that reward broad web knowledge (e.g.\ ARC-Challenge, GeoBench), which is a limitation of using a limited set of seed sources.

\subsection{Data ablations}\label{ss:controlled}

To test whether the gains come from the data rather than from the architecture, tokenizer, or training budget, we train two 600M models that are identical to {\sc Baguettotron}-600M except for their pre-training data: FineWiki (English Wikipedia, ${\sim}9$B tokens, ${\sim}17$ epochs) and FinePDFs-Edu (English, ${\sim}130$B tokens, ${\sim}1.2$ epochs).

Without post-training, neither web model produces valid multiple-choice answers. We therefore post-train both on SmolTalk~\citep{allal2025smollm2} (99,679 conversations) plus the MMLU auxiliary training split (39,852 questions, disjoint from the test set) for the answer format: ${\sim}100$M tokens over 3 epochs, without reasoning traces, adding ${\sim}0.2\%$ to their compute.

\synth{} leads by 16--17 points compared to both post-trained web models on multiple-choice and 11--14 points on open-ended tasks (\autoref{tab:controlled}, \autoref{app:benchmarks}). Both web models stay at chance on MMLU (24.0\% and 25.6\%), and dropping MMLU from the average changes it by at most 0.1 points. 

\textbf{Reasoning traces.~~} We also retrain {\sc Baguettotron}-600M on \synth{} without its reasoning traces, keeping queries and answers unchanged and matching tokens, optimizer steps, unique examples, and seed. Multiple-choice accuracy does not change (41.8\% vs.\ 42.2\% with traces), but open-ended accuracy drops from 24.3\% to 22.0\%. The loss is concentrated in truthfulness (TruthfulQA, $-9.1$) and domain reasoning (NuclearQA, $-10.0$); factual recall is unaffected ($-0.5$ on average) and ConflictQA even improves ($+3.9$; \autoref{tab:no-reasoning}).

\takeaway{Web pre-training requires a second stage, synthetic pre-training does not.} Behaviours that web models only acquire in a separate post-training stage, such as following instructions and answer formats, are written into \synth{}'s pre-training corpus. Even after the post-training stage, the web models trail by 11--17 points, so the advantage is not only one of format.

\section{Controlling factual generation with synthetic data}\label{ch:memorization}
\synth's back-translation pipeline grounds every memorization target in a real Wikipedia passage, so the training signal rewards stating verified facts rather than plausible-sounding ones. We test whether this carries over downstream by measuring factual \emph{precision} on open-ended entity descriptions (\autoref{tab:factscore}): does a model trained on grounded synthetic data confabulate less per generated fact than baselines on much larger web corpora? Verbatim copying is separately off-limits for copyright reasons~\citep{li2026combatingdatalaunderingllm}. Precision alone is not the full picture: a useful model also knows when to hedge, returning a fuzzy claim (the right century) rather than a precise but wrong one (the wrong date). \synth's reasoning traces include explicit epistemic markers for exactly this purpose, and \autoref{ss:factual-recall} tests whether \synth-trained models pick them up as a calibrated signal.

Prior synthetic augmentation does not enforce this kind of grounding. EntiGraph~\citep{ICLR2025_6dcf277e} generates entity-relation text that can introduce claims absent from the source, and Active Reading~\citep{lin2025learning} relies on self-generated strategies without an explicit fact-verification step. In both cases, teacher hallucinations propagate to the student~\citep{liu2024fictitioussyntheticdataimprove,zhu-etal-2025-enhancing}. \synth-trained models should therefore produce more factually accurate generations than models trained on either web data or unverified synthetic data.

\definecolor{ciourscol}{HTML}{1d6f3c}
\definecolor{ciwincol}{HTML}{a83232}
\definecolor{cilosecol}{HTML}{a83232}
\definecolor{cinscol}{HTML}{6b6b6b}
\providecommand{\ciboxbase}[2]{{\setlength{\fboxsep}{1.5pt}\colorbox{#1!12}{\scriptsize\textcolor{#1!70!black}{#2}}}}
\providecommand{\ciboxx}[2]{\ciboxbase{#1}{$\pm$#2}}
\providecommand{\cig}[2]{#1\,\ciboxx{ciwincol}{#2}}
\providecommand{\cir}[2]{#1\,\ciboxx{cilosecol}{#2}}
\providecommand{\cin}[2]{#1\,\ciboxx{cinscol}{#2}}
\providecommand{\cio}[2]{#1\,\ciboxx{ciourscol}{#2}}
\providecommand{\ciol}[1]{\ciboxbase{ciourscol}{#1}}
\providecommand{\cigl}[1]{\ciboxbase{ciwincol}{#1}}
\providecommand{\cirl}[1]{\ciboxbase{cilosecol}{#1}}
\providecommand{\cinl}[1]{\ciboxbase{cinscol}{#1}}

\begin{table}[t]
\centering
\caption{\textbf{Factual precision on Wikipedia entities ($n{=}500$), grouped by parameter tier.} \emph{S/(S+C)} = fraction of decomposed atomic facts \emph{Supported} by Wikipedia. \emph{macro} = per-entity Supported / total facts (Inconclusive in the denominator). \textbf{Bold} = best in tier. Pill colour encodes a paired-bootstrap test (B$=$10\,000) of each baseline against our in-tier reference: \ciol{green} $=$ ours (reference); \cigl{red} $=$ other model in-tier significantly worse compared to ours; \cinl{gray} $=$ not significant ($\alpha{=}0.05$).}
\label{tab:factscore}
\small
\setlength{\tabcolsep}{4pt}
\resizebox{\linewidth}{!}{%
\begin{tabular}{llrrrrrr}
\toprule
\textbf{Model} & \textbf{Mode} & \textbf{Tokens} & \textbf{Sup}\% & \textbf{Con}\% & \textbf{Inc}\% & \textbf{S/(S+C)} & \textbf{macro} \\
\midrule
\multicolumn{8}{l}{\textit{$\sim$50M parameters}} \\
\textbf{\textsc{Monad}} (56M, ours) & chat & 180B & 16.3 & 21.7 & 61.9 & \textbf{\cio{42.9\%}{3.8}} & \textbf{\cio{16.3\%}{1.8}} \\
\midrule
\multicolumn{8}{l}{\textit{$\sim$300--400M parameters}} \\
SmolLM2-360M-IT & chat & 4T & 26.0 & 16.0 & 58.0 & \cin{61.8\%}{3.4} & \cig{26.6\%}{2.1} \\
LFM2.5-350M & chat & 28T & 22.0 & 15.1 & 62.9 & \cig{59.3\%}{3.1} & \cig{22.0\%}{1.8} \\
Gemma-3-270M-IT & chat & 2T & 18.4 & 13.6 & 68.0 & \cig{57.6\%}{4.0} & \cig{19.4\%}{1.8} \\
OPT-350M & compl. & 180B & 4.7 & 22.5 & 72.8 & \cig{17.2\%}{4.0} & \cig{6.6\%}{1.1} \\
\textbf{\textsc{Baguettotron}-350M} (ours) & chat & 200B & 32.4 & 17.3 & 48.3 & \textbf{\cio{65.1\%}{3.2}} & \textbf{\cio{32.4\%}{2.3}} \\
\midrule
\multicolumn{8}{l}{\textit{$\sim$600M+ parameters}} \\
Qwen3-0.6B & chat & ${\sim}36$T & 31.1 & 16.3 & 52.6 & \cig{65.7\%}{4.9} & \cig{31.6\%}{2.1} \\
Phi-4-mini-instruct (3.8B) & chat & ${\sim}5$T & 27.3 & 8.0 & 64.7 & \cin{77.4\%}{3.4} & \cig{29.5\%}{1.8} \\
\textbf{\textsc{Baguettotron}-600M} (ours) & chat & 158B & 42.1 & 11.0 & 47.0 & \textbf{\cio{79.3\%}{2.1}} & \textbf{\cio{41.7\%}{2.2}} \\
\midrule
\multicolumn{8}{l}{\textit{Sparse MoEs}} \\
OLMoE-1B-7B-Instruct (7B/1B$_\text{act}$) & chat & ${\sim}5.1$T & 33.1 & 7.0 & 59.9 & \cin{82.6\%}{1.7} & \cig{33.4\%}{1.8} \\
DeepSeek-MoE-16B-Chat (16B/2.8B$_\text{act}$) & chat & ${\sim}2$T & 36.9 & 7.6 & 55.5 & \cin{82.9\%}{2.1} & \cig{38.3\%}{2.2} \\
\textbf{\textsc{Baguettotron}-MoE} (13B/1B$_\text{act}$, ours) & chat & 50B & 46.0 & 10.3 & 43.6 & \cio{81.7\%}{2.1} & \textbf{\cio{46.3\%}{2.3}} \\
\bottomrule
\end{tabular}
}
\end{table}

\subsection{Factual precision}

For each of $n{=}500$ Wikipedia seed entities sampled uniformly from {\sc Synth}'s 52{,}183-article seed corpus, we prompt the model with \emph{``What do you know about \{entity\}?''}, decompose the response into atomic facts via DeepSeek-V3.2, and label each fact \emph{Supported}, \emph{Contradicted}, or \emph{Inconclusive} against the source article, following a FActScore-inspired evaluation protocol~\cite{min2023factscore}. We report \emph{precision} $S/(S{+}C)$ on the verifiable subset, and a \emph{macro} score $S/(S{+}C{+}I)$ that penalizes inconclusive responses (\autoref{app:factscore}). Note that neither metric captures when a model chooses to abstain or hedge, which we examine in \autoref{ss:factual-recall}.

\textsc{Baguettotron}-600M and \textsc{Baguettotron}-MoE reach the highest macro scores in the table ($41.7\%$ and $46.3\%$), beating Phi-4-mini-instruct (a $6\times$ larger dense model) and DeepSeek-MoE-16B-Chat ($2.8\times$ active parameters, ${>}40\times$ more pre-training tokens). \textsc{Baguettotron}-350M ($\sim$200B SYNTH tokens) likewise outperforms LFM2.5-350M and SmolLM2-360M-IT despite seeing 10--140$\times$ fewer training tokens. The only baseline trained on a comparable token budget is OPT-350M (180B), which is also the only model that confabulates more facts than it states correctly.

\takeaway{Models trained on \synth{} are the most token-efficient factual learners in every parameter tier}, topping both precision and macro scores despite 10--140$\times$ fewer training tokens, and are the only models above $40\%$ macro on Wikipedia entities.

{
\textbf{Validation on held-out entities.~~}\label{ss:heldout} 
\synth{}'s seed determine which facts a model memorizes and we evaluate if the model states them faithfully. No model can report facts it never saw: for instance, a model trained on articles about elephants does not thereby learn about giraffes.
The evaluation prompt is not a \synth{} training query either: ``What do you know about'' occurs in none of the 5,000 queries of the released \synth{} sample.

If the model's precision came from stating safe, generic claims regardless of the entity, it would persist outside the training data. We test this on 600 Wikipedia Good Articles that are not Vital articles and are absent from the seed corpus, under the same protocol (\autoref{app:heldout}). On held-out entities, {\sc Baguettotron}-MoE abstains on 67\% (vs.\ 20\% in-seed), and its precision on attempted answers drops from 82\% to 62\%. The metric therefore tracks what the model knows. OLMoE-1B-7B-Instruct~\citep{muennighoff2025olmoe}, a web-trained MoE of the same active size, abstains on only 7\% and keeps 81\% precision. As web-scale data covers more of these entities, this is expected. 
}

\subsection{Calibrating factual precision with epistemic markers}\label{ss:factual-recall}

\synth's memorization traces include explicit epistemic markers that flag individual claims as confident or uncertain. We test whether models trained on \synth{} learn to deploy these markers in a calibrated way, so whether traces dominated by uncertainty markers correspond to less-factual outputs, and whether the model abstains by committing to fewer claims. For each trace, we count high-confidence and low-confidence symbols, then bucket the trace as \emph{Confident} (high markers dominate) or \emph{Uncertain} (low markers dominate); ties are dropped.\footnote{FActScore's atomic-fact extractor occasionally produces ``meta-claims'' about the speaker (e.g., ``the speaker is afraid'') from refusal preambles, ${\sim}6\%$ of all extracted atomic claims, predominantly labeled \emph{Inconclusive}. These slightly compress the FActScore gaps reported here.}

\begin{figure}[t]
    \centering
    \begin{subfigure}[t]{0.49\textwidth}
        \includegraphics[width=\textwidth]{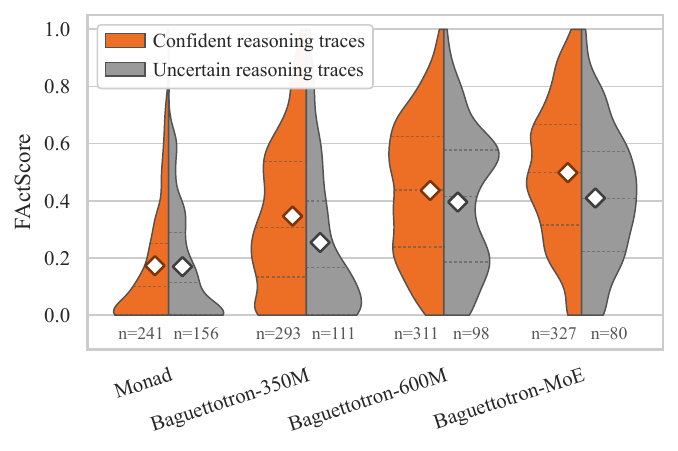}
        \subcaption{Per-topic FActScore distribution.}
        \label{fig:fp-factscore}
    \end{subfigure}
    \hfill
    \begin{subfigure}[t]{0.49\textwidth}
        \includegraphics[width=\textwidth]{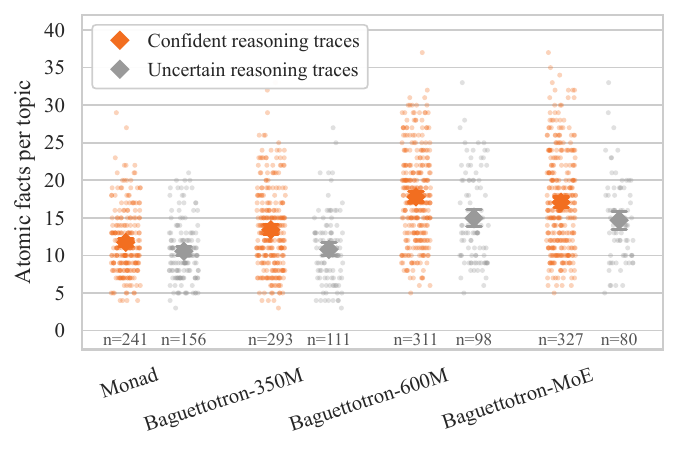}
        \subcaption{Atomic facts extracted per topic.}
        \label{fig:fp-factcount}
    \end{subfigure}
    \caption{\textbf{Epistemic markers in reasoning traces predict factual quality and output volume.} Per-topic FActScore (a) and atomic-fact count (b), conditioned on whether the trace is dominated by confident or uncertain markers (majority count; ties dropped). Diamonds mark conditional means.}
    \label{fig:factual-precision-calibration}
\end{figure}

\autoref{fig:fp-factscore} shows that three of four models calibrate factual precision against the markers: \textsc{Baguettotron}-350M ($0.35{\to}0.25$), \textsc{Balanced}-600M ($0.44{\to}0.40$), and \textsc{Baguettotron}-MoE ($0.50{\to}0.41$) all drop FActScore on uncertain-tagged traces, with the larger models showing the cleanest separation. Only for \textsc{Monad} (56M), the markers carry no signal ($0.17{\to}0.17$).

\autoref{fig:fp-factcount} shows that all four models calibrate output volume: the 350M, 600M, and MoE variants commit to ${\sim}15{-}20\%$ fewer atomic facts on uncertain traces, and even \textsc{Monad} shrinks from $11.7$ to $10.6$ facts per topic. While modest, \textsc{Monad} learns to \emph{say less} under uncertainty without learning to \emph{be more right}, suggesting that calibrated abstention is acquired earlier in the parameter-count curve than calibrated precision. For \synth{}-trained models above ${\sim}300\text{M}$, the epistemic markers function as a usable confidence indicator for both output volume and factual precision.

\takeaway{Models trained on \synth{} learn to use its epistemic markers as a
signal}. The uncertainty markers predict both lower factual precision (350M and above) and reduced output volume (all sizes).

\subsection{Domain and task adaptation}\label{ss:tele}

A key advantage of synthetic generation is the ability to target specific domains and skills required of a model. We illustrate this through adaptation to telecommunications, a domain characterized by scarce specialized data and dense terminology. Our seed corpus combines two complementary sources: a telecom-focused slice of English Wikipedia ($\sim 3$M tokens) and a corpus of 3GPP technical standards~\citep{maatouk2024telellmsseriesspecializedlarge} ($\sim 90$M tokens). Applying the same synthetic generation pipeline to this seed material with task-targeted prompts (memorization, mcq, and free-form QA), we produce $\sim$460M synthetic tokens ($10\times$ paragraph amplification). Fine-tuning the 600M base model on this yields consistent gains across both public and internal benchmarks: TeleQnA accuracy improves from 41.6\% to 56.7\% (+15.1\%), and FactScore on 3GPP standards from 21.5\% to 38.8\% (+17.3\%). 

On LLM-as-a-judge evaluation (Claude Opus 4.5, 50 telecom questions scored on 0-100 across accuracy, hallucination, reasoning, and answer quality), the 3GPP-augmented model reaches 67.8, a 15.4\% improvement over a wiki-only telco fine-tune (58.7) and $2.4\times$ the score of the 600M baseline (27.8). The gains come mostly from precise domain knowledge (\emph{e.g.} 5G) and the disambiguation of acronyms, where the baseline tends to hallucinate plausible but incorrect expansions.

\textbf{Tool calling.~~}\label{ss:tool-calling} We also adapt the base models to tool calling, generating the data with the same two-auxiliary design as \synth{}. A \emph{drafter} writes a user query, synthetic tool schemas, and the gold call from one seed paragraph and seven sampled constraints (environment, query type, single or multiple calls, menu size up to 25, positive or negative outcome, language, style); a \emph{solver} writes the reasoning trace. Both are fine-tuned on Gemma-3-12B-base, and ten passes over 28,428 seed passages yield 532k rows (578M tokens). 
We continue training the base models on a 70/30 mix of this data and general \synth{} for 5,700 steps (${\sim}1.5$B tokens), with the same next-token objective and no RL. 
On BFCL v2~\citep{patil2025bfcl} (full split, $n{=}3{,}981$; checkpoint selected on a 1,197-item dev subset), {\sc Baguettotron}-600M reaches 53.1\%, 3.9 points above FunctionGemma-270M (7.8 on macro accuracy) on ${\sim}40\times$ fewer training tokens (\autoref{tab:tool-calling}).
The base model emits no tool calls, so the capability comes from the tool-calling data. As this data shares the format and objective of \synth{}, it can be folded into the pre-training mix, which we leave to future work.

\takeaway{Synthetic generation enables targeted domain adaptation from small seed corpora.} A modest in-domain seed ($\sim$93M tokens), amplified $10\times$ through task-targeted synthesis, more than doubles judge scores over the base 600M model and substantially improves factual grounding.

\section{Conclusion}
We have introduced \synth, the first open synthetic-only pretraining corpus designed to support the entire training curriculum as a single stage. The \textsc{Baguettotron} suite, trained from scratch on \synth\ across dense (56M to 600M parameters) and Mixture-of-Experts (13B/1B active) configurations, undergoes no separate SFT or RLHF stage yet follows instructions, reasons, and recalls facts as native capabilities of a single training stage. The same 600M model pre-trained on web data produces no valid answers without post-training, and still trails by double digits after it. Training without reasoning traces costs up to 10 points on truthfulness and domain reasoning. Our models reach the highest FActScore in every parameter tier on 10--140$\times$ fewer tokens, and recognize the limits of their knowledge: outside the seeds, {\sc Baguettotron}-MoE abstains on 67\% of entities, against 7\% for a web-trained MoE. Beyond pre-training, synthetic data adapts the base model to new domains.

These results recontextualize the scaling literature. The Chinchilla coefficients were fit on web text~\citep{hoffman2022scaling}, which is weakly aligned with the capabilities a small model needs to learn. Engineered data shifts the curve as training remains productive well past the canonical $20\times$ token-to-parameter ratio, and capacity-bound predictions from controlled-synthetic experiments transfer cleanly to deployable generalist models (\autoref{ch:memorization}). This confirms the initial intuition from the Phi series~\citep{li_textbooks_2023} that data quality and learnability is an additional controllable axis alongside model size and token count and has been insufficiently optimized by LLM research.

Finally, \synth{} opens up new paths for data releasability and transparency. We showed that a viable pretraining environment could be built from a small collection of 50,000 Wikipedia articles. Conversely, open data sources have consistently emerged as higher quality sources than webcrawl and better suited for large-scale seed infrastructures: Wikipedia and Wikidata are already integral components of synthetic pipelines for generalist and search-specialized models~\citep{huang_step_2026,li_websailor_2025,bashir_chroma_2026}. We already illustrated this with tool-calling where, with synthetic data, a 600M base model outperforms FunctionGemma-270M, trained on 6T tokens.

\section{Limitations and future work}
Our current research showed that our synthetic training recipe is inherently scalable: models in a larger size range assimilate more facts and do it faster. Yet, to move beyond the current category of models, our pipelines will also need to scale across multiple axes.

\textbf{Seed coverage and infrastructure.~~} Our current design is bounded by the information contained in the $\sim$58k Wikipedia seed articles. While this ensures tight experimental control over what the model learns, it imposes a hard ceiling on parametric knowledge. Wikipedia is a natural expansion space, especially across different multilingual versions, and has been shown to reliably improve on advanced knowledge benchmarks~\citep{lin2025learning}.

\textbf{Cultural diversity.~~} Our current seeding infrastructure intently selects knowledge relevant to the English-speaking Wikipedia contributor, which aligns strongly with standard LLM benchmarks. Our multilingual pipeline covers multiple languages but was shaped by English seeds.
Future work should extend seed selection to native language content.

\textbf{Capabilities.~~} The exercise difficulty in current \synth{} is currently calibrated for SLMs use cases. Scaling this pipeline to larger, more generalist model calls for a new synthetic pipeline, extending to code generation, synthesized agentic scenarios (interleaved thinking, multi-step tool use, environment-grounded trajectories) and more broadly long horizon tasks.

\textbf{Hybrid pretraining mixes.~~} Our current setting isolates the contribution of the synthetic corpus by removing all confounding from web data. Yet it also requires engineering a wide range of model behaviors (including refusal) that may be captured at scale through organic data training.

\section*{Acknowledgements}

We acknowledge the EuroHPC Joint Undertaking for awarding this project access to the EuroHPC supercomputer MareNostrum~5, hosted by the Barcelona Supercomputing Center (BSC), through the EuroHPC Extreme Scale Access call (EHPC-EXT-2025E01-092, \emph{JULIP: Powerful LLMs made in Europe}).
This work was granted access to the HPC resources of IDRIS under the allocations A0191016886 (`\emph{EuroSynth}') and AD011014736R1 made by GENCI. Parts of this research received funding from SPRIN-D, the German Federal Agency for Breakthrough Innovation.
We also thank Oleg Filatov, Vedant Nanda and Jiangtao Wang for their helpful feedback on this project.

\bibliographystyle{plainnat}   
\bibliography{papers}       

\appendix

\section{Query generator constraint priors}\label{app:query-constraints}

The query generator (\autoref{ss:memorization}) samples constraints per call from independent priors over six axes:
\begin{itemize}
    \item \emph{Query type}: information retrieval, problem-solving, analytical, comparative, predictive, integrative.
    \item \emph{Complexity}: simple, moderate, complex.
    \item \emph{User profile}: expert, professional, informal reader.
    \item \emph{Query result}: positive ($0.80$), negative ($0.10$), absurd ($0.05$), ambiguous ($0.05$).
    \item \emph{Target language}: drawn from the deployed language set (\autoref{fig:words-lang}).
    \item \emph{Query style} (added in a more recent subset): direct, indirect, oral.
\end{itemize}
Unless noted, axes use a uniform prior over their listed values; the \emph{query result} priors are the only deliberately skewed ones, motivated in \autoref{ss:memorization}.

\section{Training details}\label{app:training-runs}

\paragraph{\textsc{Monad}-56M.} Trained with Nanotron in approximately $11$ hours on 16 H100s over ${\sim}180$B tokens. Sequence length $1{,}024$, global batch size $1{,}024$ sequences (local $32$, grad-accum $2$, ${\sim}1.05$M tokens/step), peak lr $3{\times}10^{-3}$ with $10{,}000$-step warmup, $150{,}000$ steps; followed by a $10{,}000$-step context-extension phase at sequence length $2{,}048$ (local $16$, grad-accum $4$, ${\sim}2.10$M tokens/step, peak lr $1{\times}10^{-3}$, $1{,}000$-step warmup). The custom 8k-vocabulary tokenizer was trained directly on the English segment of {\sc Synth}. Steady-state MFU $13$--$17\%$ (the lower end at seq=$1024$ rising to the upper end during the seq=$2048$ extension), final cross-entropy $1.52$.

\paragraph{\textsc{Baguettotron}-350M.} Deep-narrow architecture ($d{=}576$, 80 layers, 321M parameters) trained with Nanotron over ${\sim}199$B tokens. $170{,}000$ total steps over ${\sim}54$ hours on 16 H100s, sequence length $2{,}048$, global batch size $512$ sequences (local $8$, grad-accum $4$, ${\sim}1.05$M tokens/step), peak lr $3{\times}10^{-3}$ with $10{,}000$-step warmup; followed by a $20{,}000$-step context-extension phase at sequence length $4{,}096$ (local $4$, grad-accum $8$, ${\sim}2.10$M tokens/step, peak lr $1{\times}10^{-3}$, $2{,}000$-step warmup). Steady-state MFU $\sim$$25\%$ in the production run (the deep $d{=}576$ ablation at seq=$4096$ sits at $17$--$19$\% MFU), final cross-entropy $1.15$.

\paragraph{\textsc{Baguettotron}-600M.} $151{,}000$ steps over ${\sim}52.6$ hours across two SLURM jobs at global batch size $512$ (local $8$, ${\sim}1.05$M tokens/step), peak lr $1.5{\times}10^{-3}$ with $10{,}000$-step warmup, $158.3$B tokens, ${\sim}300$\,TFLOPS/GPU steady-state (${\sim}30\%$ MFU). The architecture-ablation runs at sequence length $4096$ sit closer to $21$--$31\%$ MFU depending on hidden dimension.

\paragraph{\textsc{Baguettotron}-MoE.} $31{,}847$ steps at global batch size $768$ (local $48$, ${\sim}1.57$M tokens/step), peak lr $3{\times}10^{-3}$ with $1{,}000$-step warmup, ${\sim}50$B tokens (less than one pass through {\sc Synth}). The load-balance auxiliary loss ($10^{-3}$) is the only addition over the dense recipe. Steady-state MFU is ${\sim}17.5\%$, and final cross-entropy is $1.19$ at step $31{,}840$, matching dense {\sc Baguettotron}-600M while seeing ${\sim}32\%$ as many tokens. Checkpoints are written every $636$ steps (${\sim}1$B tokens).

\section{Compute accounting}\label{app:compute}

\paragraph{Cluster compute.} We collected every SLURM job related to \synth{}, including development runs (\autoref{tab:compute}). All jobs ran on Nvidia H100s, 4 per node. Producing the final corpus took 6,144.6 of the 16,044 GPU-hours; the rest is exploratory development.

We did not track MFU for every job. At an assumed 20\% MFU against the H100 SXM dense BF16 peak (989.4 TFLOP/s), final corpus production amounts to ${\sim}4.4{\times}10^{21}$ FLOPs, $164\times$ below the $7.2{\times}10^{23}$ reported for Llama~3~8B~\citep{grattafiori2024llama3}.

\begin{table}[h]
\centering
\small
\caption{\textbf{Cluster compute.} GPU-hours of all \synth{}-related jobs (H100), including development. Excludes the FineWiki and FinePDFs-Edu 600M runs (broken down in \autoref{tab:compute-models}).}
\label{tab:compute}
\begin{tabular}{lr}
\toprule
\textbf{Category} & \textbf{GPU-hours} \\
\midrule
Inference (generation) & 10,804 \\
Training & 4,936 \\
Fine-tuning & 195 \\
Evaluation & 71 \\
Embeddings & 39 \\
\midrule
\textbf{Total} & \textbf{16,044} \\
\bottomrule
\end{tabular}
\end{table}

\paragraph{Frontier-model supervision.} The GPU-hours exclude the Gemini~2.5~Pro~\citep{comanici2025gemini25} calls that produce the Stage-1 training sets for the auxiliaries (\autoref{tab:stage1}).

\begin{table}[h]
\centering
\small
\caption{\textbf{Stage-1 frontier-model supervision} (Gemini~2.5~Pro) per auxiliary task (\autoref{ch:synth}).}
\label{tab:stage1}
\begin{tabular}{lrr}
\toprule
\textbf{Stage-1 category} & \textbf{Calls} & \textbf{Output tokens} \\
\midrule
Query model & 7,267 & 13.3M \\
Memorization & 6,006 & 10.2M \\
RAG & 7,108 & 23.2M \\
Arithmetic & 6,927 & 6.9M \\
Creative writing & 14,584 & 29.6M \\
Editing & 15,963 & 27.4M \\
MCQ & 20,477 & 21.2M \\
Conversation & 2,932 & 1.4M \\
\midrule
\textbf{Total} & \textbf{81,264} & \textbf{133.2M} \\
\bottomrule
\end{tabular}
\end{table}

\paragraph{Per-model cost.} We amortize the final corpus production over the four models trained on it in this paper (\autoref{tab:compute-models}). The generation share shrinks with every further model trained on the released corpus.

\begin{table}[h]
\centering
\small
\caption{\textbf{End-to-end GPU-hours of the controlled 600M runs} (\autoref{ss:controlled}), excluding Stage-1 frontier supervision.}
\label{tab:compute-models}
\begin{tabular}{lrrrr}
\toprule
\textbf{Model} & \textbf{Pre-training} & \textbf{Post-training} & \textbf{Generation share} & \textbf{Total} \\
\midrule
\textsc{Baguettotron}-600M (\synth{}) & 843 & -- & 1,536 & 2,379 \\
FineWiki 600M & 941 & ${\sim}4$ & -- & ${\sim}945$ \\
FinePDFs-Edu 600M & 1,015 & ${\sim}4$ & -- & ${\sim}1{,}019$ \\
\bottomrule
\end{tabular}
\end{table}

\section{Safety and integrity evaluation}\label{app:propella}

\begin{figure}[h]
    \centering
    \includegraphics[width=\linewidth]{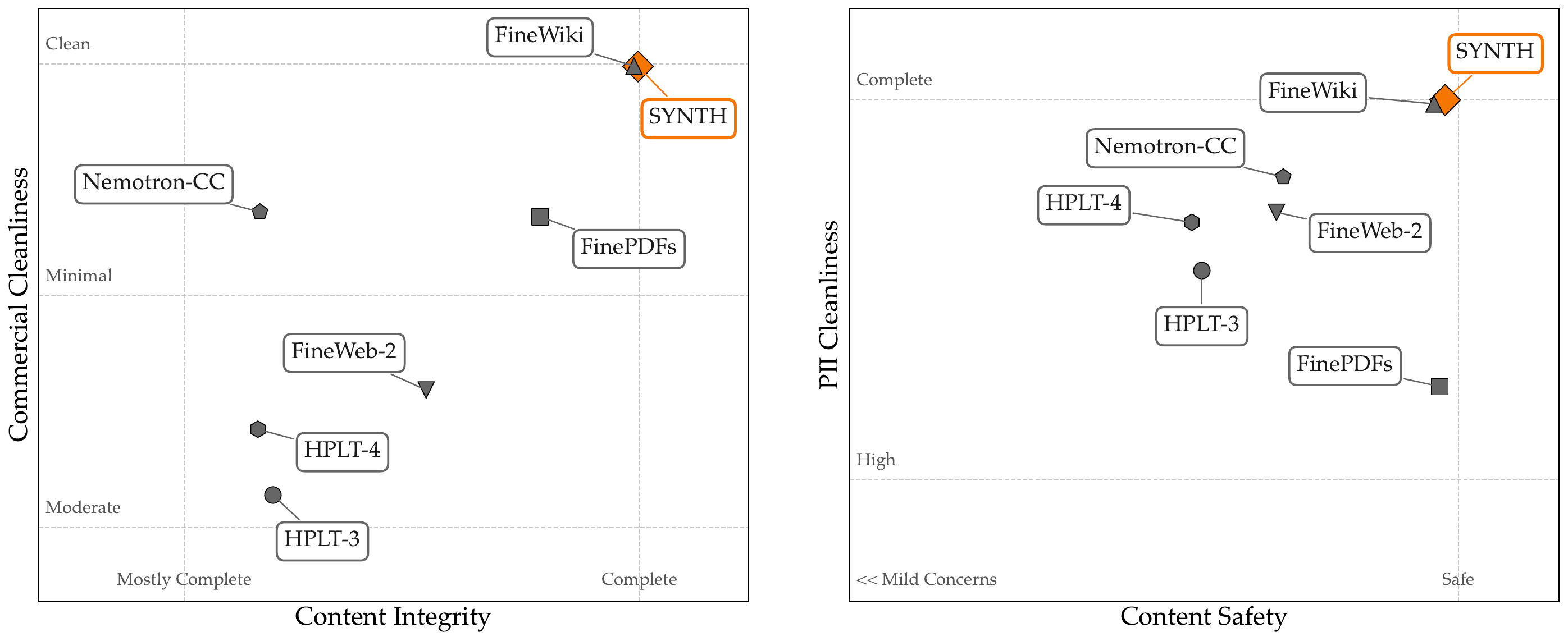}
    \caption{Propella evaluation on safety and integrity axes.}
    \label{fig:propella-bottom}
\end{figure}

In Figure~\ref{fig:propella-bottom}, we show that \synth{} is located in the top-right corner of the integrity/safety axes from the \texttt{propella-1} model, naturally matched only by the Wikipedia-sourced FineWiki dataset. Combined with the results in Figure~\ref{fig:propella}, \synth{} is the only open pre-training corpus located in the top-right corner along all evaluated axes. 

\section{Second-classifier validation of data quality (Propella)}\label{app:second-judge}

To check that the Propella-1 scores do not simply reflect annotator alignment with encyclopedic style, we score six English corpora with \texttt{HuggingFaceFW/fineweb-edu-classifier}~\citep{penedo2024the}.
This forms a second, independent evaluation of data quality (\autoref{tab:second-judge}).

 We  evaluate 100 English documents per corpus. The classifier reads 512 tokens, so we average its scores over up to four 512-token windows covering the first 2,048 tokens. \synth{} documents are scored in their pre-training format (query, reasoning, answer). Nemotron-CC is represented by the high-quality bucket of a public 1M-document sample.

\begin{table}[h]
\centering
\small
\caption{\textbf{Educational value by an independent classifier} (\texttt{fineweb-edu-classifier}, scale 0--5, higher is better), 100 English documents per corpus, with 95\% CIs.}
\label{tab:second-judge}
\begin{tabular}{lr}
\toprule
\textbf{Corpus} & \textbf{Mean score} \\
\midrule
\synth{} & \textbf{2.23} {\scriptsize [2.10, 2.35]} \\
Nemotron-CC & 1.72 {\scriptsize [1.56, 1.88]} \\
FineWiki & 1.64 {\scriptsize [1.49, 1.79]} \\
Common Corpus & 1.49 {\scriptsize [1.36, 1.62]} \\
FinePDFs & 1.34 {\scriptsize [1.19, 1.49]} \\
FineWeb & 1.21 {\scriptsize [1.09, 1.33]} \\
\bottomrule
\end{tabular}
\end{table}

\synth{} scores highest, and the ordering agrees with Propella-1's educational-value ranking. FineWiki, the most encyclopedic corpus, is not favoured.

\section{FActScore evaluation protocol}\label{app:factscore}
We evaluate factual precision by prompting models with ``what do you know about [entity].'' for entities sampled from {\sc Synth}'s 52,183 Wikipedia seed articles. The model's response (excluding any reasoning trace) is decomposed into atomic facts by a strong judge model (DeepSeek-V3, fp8), and each fact is verified against the corresponding Wikipedia source article

For example, for the entity \emph{Scipio Africanus}:

\begin{enumerate}
    \item \textbf{Prompt:} ``What do you know about Scipio Africanus?''
    \item \textbf{Model output (content only, reasoning trace excluded):} ``Publius Cornelius Scipio Africanus Major was a Roman general and statesman who rose to prominence during the Second Punic War\ldots''
    \item \textbf{Atomic fact decomposition:} \texttt{["Scipio Africanus was a Roman general", "Scipio Africanus was a Roman statesman", "Scipio Africanus lived from 236 to 183 BCE", \ldots]}
    \item \textbf{Verification against source passage:} each fact is classified as \emph{Supported}, \emph{Contradicted}, or \emph{Inconclusive} by the judge model given the Wikipedia article text
    \item \textbf{FActScore} = \#Supported / \#total atomic facts
\end{enumerate}

Note that we evaluate the model's \emph{content} output only, not its internal reasoning trace. The number of atomic facts per response varies (typically 5--25); models that generate more specific claims are exposed to more verification opportunities. We report both the score and the average number of facts per response.

\textbf{Statistics.} Confidence intervals are computed by cluster bootstrap with $10{,}000$ resamples over the $n{=}500$ entities (per-fact resampling would inflate effective sample size since facts are correlated within entity). Pairwise comparisons use the \emph{paired Wilcoxon signed-rank test} on per-entity precision, dropping entities where either model has $S{+}C{=}0$.

\section{Held-out entity evaluation}\label{app:heldout}

This appendix details the held-out validation summarized in \autoref{ss:heldout}.

\paragraph{Construction.} We start from English Wikipedia Good Articles, which are long enough to verify atomic facts against, and remove all Vital articles (levels 1--5) and all seed articles. Removing Vital articles avoids topics adjacent to the seeds, and leaves 41,927 candidates with no seed overlap. From this pool we sample 600 entities, stratified to match the 500 in-seed entities on $\log_{10}$ daily pageviews over 60 days (median 213 vs.\ 186) and on entity type (26\% persons). The prompt and FActScore protocol are those of \autoref{app:factscore}, with the full article extract as reference. As a baseline, we use OLMoE-1B-7B-Instruct~\citep{muennighoff2025olmoe}, the web-trained MoE with the same active size in \autoref{tab:factscore}.

\paragraph{Abstention and metric.} DeepSeek-V4-Pro labels each response as \emph{abstain} or \emph{attempt}. It agrees with a refusal regex on 90.5\% of responses and gives the same aggregate rates; most disagreements are confabulations that contain a stray disclaimer. Abstaining responses still decompose into atomic claims, mostly meta-claims about the model's own knowledge, which inflate S/(S+C) (\autoref{tab:heldout-coverage}). \autoref{tab:heldout} therefore reports precision over attempted responses only. The in-seed and held-out tiers also differ in reference (seed passages vs.\ full articles) and judge version, but OLMoE's stable precision (82.7\% vs.\ 81.1\%) indicates that this shift does not bias S/(S+C).

\begin{table}[h]
\centering
\small
\caption{\textbf{Abstention and precision on held-out entities.} Abstention rate (LLM judge) and S/(S+C) over attempted responses, with 95\% CIs (Wilson for abstention; entity-level bootstrap, $B{=}10{,}000$, for precision). \emph{No title mention}: held-out entities whose title never occurs in the \synth{} training text.}
\label{tab:heldout}
\begin{tabular}{llrrr}
\toprule
\textbf{Model} & \textbf{Entities} & \textbf{$n$} & \textbf{Abstain (\%)} & \textbf{S/(S+C), attempted (\%)} \\
\midrule
\textsc{Baguettotron}-MoE & in-seed & 500 & 19.8 {\scriptsize [16.5, 23.5]} & 81.8 {\scriptsize [79.5, 84.0]} \\
 & held-out & 600 & 66.8 {\scriptsize [63.0, 70.5]} & 61.7 {\scriptsize [56.2, 66.8]} \\
 & \quad no title mention & 302 & 74.2 {\scriptsize [69.0, 78.8]} & 55.2 {\scriptsize [46.1, 63.9]} \\
\midrule
OLMoE-1B-7B-Instruct & in-seed & 500 & 0.8 {\scriptsize [0.3, 2.0]} & 82.7 {\scriptsize [81.0, 84.4]} \\
 & held-out & 600 & 7.2 {\scriptsize [5.4, 9.5]} & 81.1 {\scriptsize [79.6, 82.6]} \\
 & \quad no title mention & 302 & 11.9 {\scriptsize [8.7, 16.1]} & 77.5 {\scriptsize [75.0, 79.8]} \\
\bottomrule
\end{tabular}
\end{table}

\paragraph{Results.} Outside its seeds, {\sc Baguettotron}-MoE abstains on 67\% of entities, against 20\% in-seed, while OLMoE abstains on only 7\% (\autoref{tab:heldout}). When our model does answer, its precision drops from 82\% to 62\%, so \synth{} gives faithful recall of its seeds rather than broader coverage. OLMoE keeps ${\sim}81\%$ precision, as web-scale pre-training covers many of these etities.

\begin{table}[h]
\centering
\small
\caption{\textbf{Atomic-fact counts on the 600 held-out entities}, split by the judge's abstain/attempt label. $S$, $C$, $I$: Supported, Contradicted, Inconclusive facts. 95\% CIs by entity-level bootstrap ($B{=}10{,}000$).}
\label{tab:heldout-coverage}
\begin{tabular}{llrrrrr}
\toprule
\textbf{Model} & \textbf{Responses} & \textbf{$n$} & \textbf{$S$} & \textbf{$C$} & \textbf{$I$} & \textbf{S/(S+C) (\%)} \\
\midrule
\textsc{Baguettotron}-MoE & attempt & 199 & 960 & 596 & 1,944 & 61.7 {\scriptsize [56.2, 66.8]} \\
 & abstain & 401 & 1,988 & 299 & 2,429 & 86.9 {\scriptsize [85.0, 88.7]} \\
 & total & 600 & 2,948 & 895 & 4,373 & 76.7 {\scriptsize [74.0, 79.3]} \\
\midrule
OLMoE-1B-7B-Instruct & attempt & 557 & 10,581 & 2,459 & 11,763 & 81.1 {\scriptsize [79.6, 82.6]} \\
 & abstain & 43 & 401 & 87 & 559 & 82.2 {\scriptsize [77.3, 86.5]} \\
 & total & 600 & 10,982 & 2,546 & 12,322 & 81.2 {\scriptsize [79.7, 82.6]} \\
\bottomrule
\end{tabular}
\end{table}

\paragraph{Leakage.} Retrieval in \synth{} runs only over the seed articles, and the Good Articles were collected for this experiment only, so no held-out passage can reach the corpus. Titles can still appear incidentally, as Wikipedia is densely cross-referenced. We therefore scan the assembled training text of all 77,908,583 \synth{} rows with Aho--Corasick exact matching (case-sensitive, word-boundary filtered, patterns of at least 5 characters). Of the 600 titles, 298 occur at least once (279,016 occurrences, median 114 per matched entity) and 302 never do; adding MediaWiki redirect aliases raises the matched set to 360 (390,921 occurrences), leaving 240 clean. These counts overstate exposure: matches carry no facts, and the most frequent are generic strings (\emph{The 1}: 87,597; \emph{5,6,7,8}: 16,655) or parts of other names (\emph{Mario Bros.}). On the 302 title-clean entities, the abstention gap widens to 74\% vs.\ 12\% (\autoref{tab:heldout}), and to 76.7\% vs.\ 14.6\% on the 240 alias-clean ones. Both models abstain less on mentioned entities ({\sc Baguettotron}-MoE 59\%, OLMoE 2\%), so mentions also track general familiarity.

\section{Benchmark suite: full results}\label{app:benchmarks}
\paragraph{Models.} Four {\sc Synth} based models (56M, 350M, 600M dense, 13B/1B-active MoE) and four open small models in the 270M--600M parameter range: Gemma-3-270M, SmolLM2-360M, LFM2.5-350M, Qwen3-0.6B. Baselines see 2--36T tokens of curated web data; {\sc Baguettotron} sees 50--200B tokens of {\sc Synth}.

\begin{figure}[h]
\centering
\includegraphics[width=\linewidth]{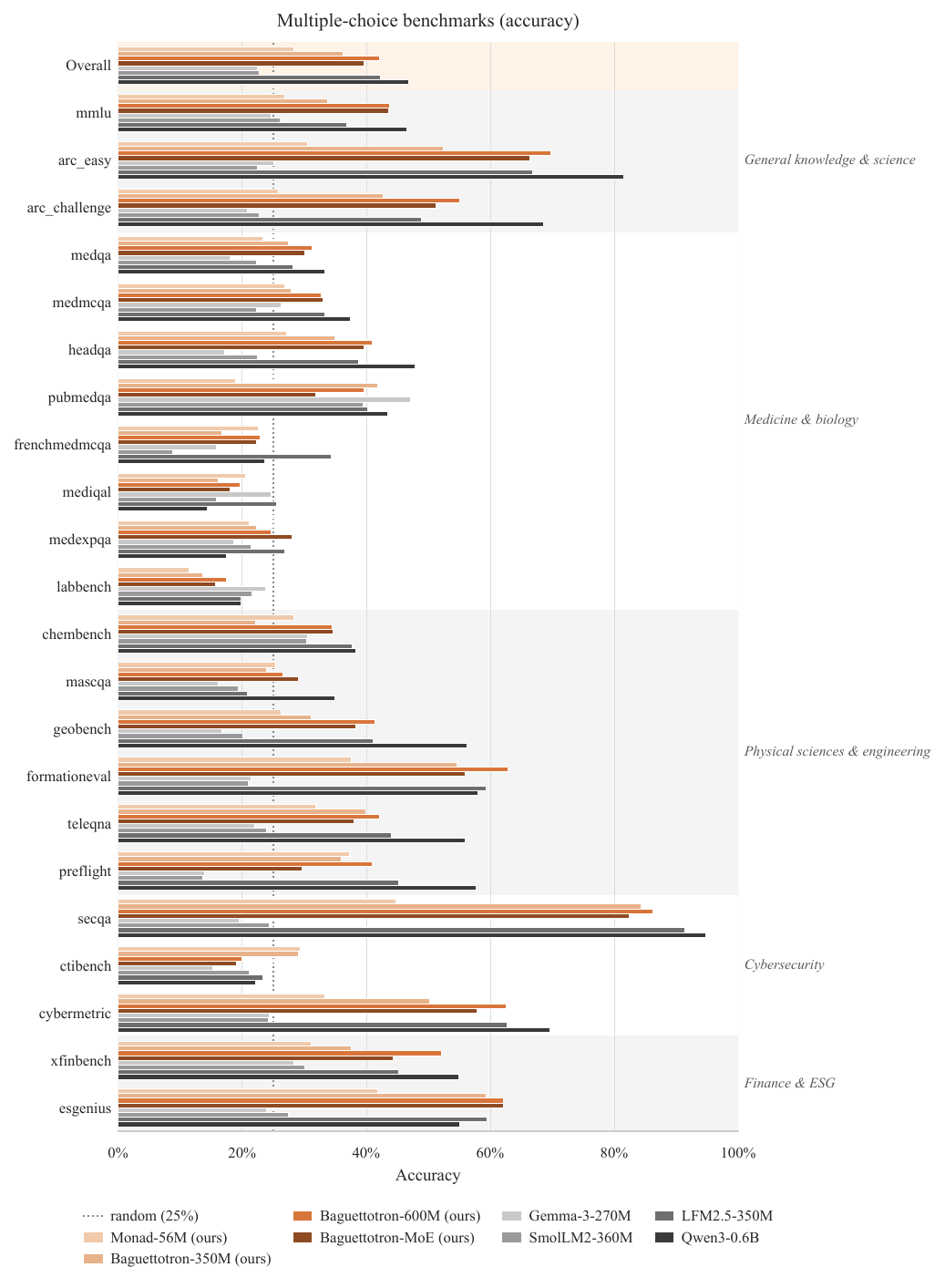}
\caption{Per-benchmark accuracy on the 22 multiple-choice tasks. Dotted line: 4-way random baseline. The \emph{Overall} row (highlighted) is the simple average across the 22 tasks. {Tasks are ordered by knowledge domain (right margin).}}
\label{fig:benchmarks-mcq}
\end{figure}

\begin{figure}[h]
\centering
\includegraphics[width=\linewidth]{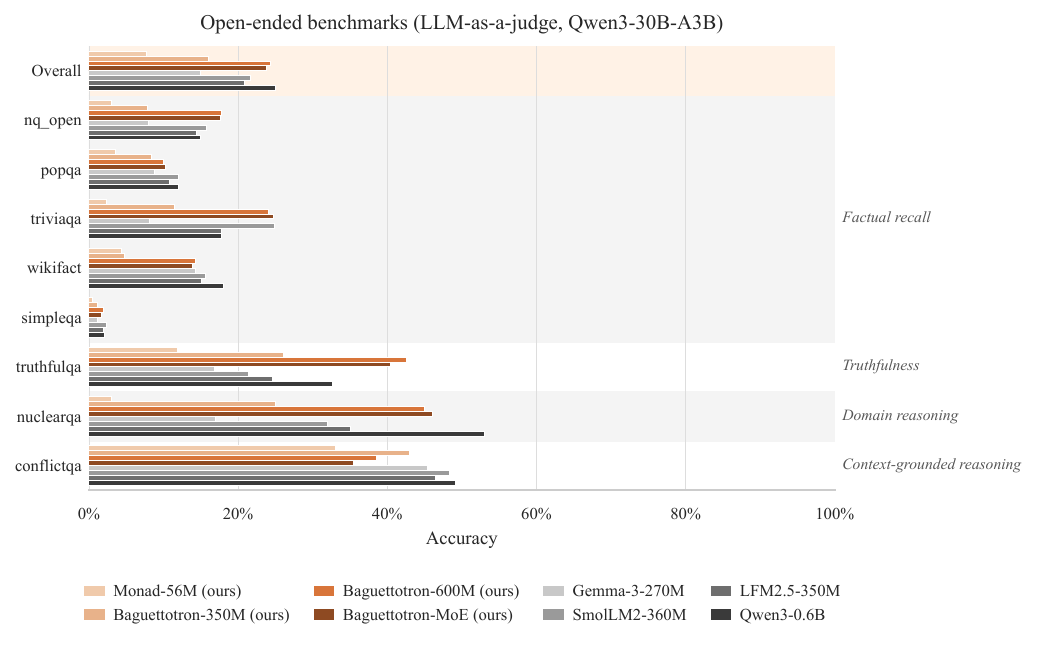}
\caption{Per-benchmark CORRECT answers on the 8 open-ended tasks. {Tasks are ordered by capability group (right margin).}}
\label{fig:benchmarks-oe}
\end{figure}

\paragraph{Benchmarks.} 22 multiple-choice tasks (MMLU, ARC-Easy, ARC-Challenge, plus vertical benchmarks spanning medicine, science, finance, cyber, geography and engineering) and 8 open-ended QA tasks (closed-book recall, reading comprehension, calibration). The full list is given on the y-axes of \autoref{fig:benchmarks-mcq} and \autoref{fig:benchmarks-oe}. Multilingual benchmarks are evaluated over \emph{all} available language splits, not English alone, with per-task scores averaged across splits; this exercises the non-English capabilities induced by \synth's multilingual generation (\autoref{fig:words-lang}).

\begin{figure}[h]
\centering
\includegraphics[width=\linewidth]{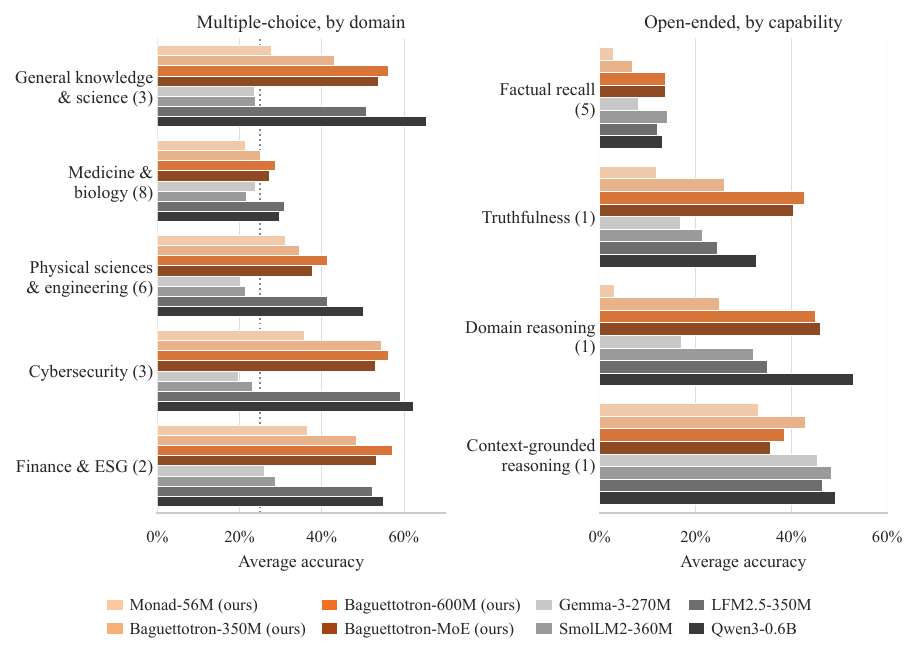}
\caption{Average accuracy per grouped capability. Multiple-choice tasks are grouped by knowledge domain, open-ended tasks by the capability they probe; the number of tasks per group is in parentheses. Dotted line: 4-way random baseline.}
\label{fig:benchmarks-groups}
\end{figure}

\begin{table}[h]
\centering
\caption{\textbf{Pre-training data ablation at 600M.} Identical architecture, tokenizer, and training steps. Only the \synth{} model is evaluated without post-training. \emph{w/o MMLU}: 21-task average, since the MMLU training split is in the web models' post-training mix.}
\label{tab:controlled}
\small
\begin{tabular}{llrrr}
\toprule
\textbf{Pre-training data} & \textbf{Post-training} & \textbf{MCQ (22)} & \textbf{MCQ w/o MMLU (21)} & \textbf{Open-ended (8)} \\
\midrule
FineWiki & SmolTalk + MMLU-aux & 26.1 & 26.2 & 10.2 \\
FinePDFs-Edu & SmolTalk + MMLU-aux & 25.1 & 25.1 & 13.5 \\
\synth{} & none & \textbf{42.2} & \textbf{42.2} & \textbf{24.3} \\
\bottomrule
\end{tabular}
\end{table}

\begin{table}[h]
\centering
\small
\caption{\textbf{Reasoning-trace ablation.} {\sc Baguettotron}-600M trained on \synth{} with and without reasoning traces (\autoref{ss:controlled}), grouped by capability. Matched on tokens, optimizer steps, unique examples, and seed; one run each.}
\label{tab:no-reasoning}
\begin{tabular}{llrrr}
\toprule
\textbf{Group} & \textbf{Benchmark} & \textbf{With traces} & \textbf{Without} & \textbf{$\Delta$} \\
\midrule
Factual recall & NQ-Open & 17.7 & 16.6 & $+1.1$ \\
 & PopQA & 10.0 & 9.8 & $+0.2$ \\
 & TriviaQA & 24.0 & 25.3 & $-1.3$ \\
 & WikiFact & 14.3 & 10.9 & $+3.4$ \\
 & SimpleQA & 1.9 & 2.7 & $-0.8$ \\
Truthfulness & TruthfulQA & 42.6 & 33.5 & $+9.1$ \\
Domain reasoning & NuclearQA & 45.0 & 35.0 & $+10.0$ \\
Context-grounded reasoning & ConflictQA & 38.5 & 42.4 & $-3.9$ \\
\midrule
\multicolumn{2}{l}{Open-ended overall (8)} & \textbf{24.3} & 22.0 & $+2.3$ \\
\multicolumn{2}{l}{Multiple-choice overall (22)} & \textbf{42.2} & 41.8 & $+0.4$ \\
\bottomrule
\end{tabular}
\end{table}

\paragraph{Protocol.} All evaluations are zero-shot and served with vLLM. Every model is run with \texttt{chatml}-format chat templating ({\sc Gemma} uses its native \texttt{<start\_of\_turn>} markers): the user turn carries the question and the lettered choices for MCQ, generation begins after \texttt{<|im\_start|>assistant}.

{\sc PleIAs} models ({\sc Baguettotron-300M}, {\sc balanced-600M}, {\sc MoE}, {\sc Monad}) are {\sc SYNTH}-trained thinking models, evaluated without a system prompt and with \texttt{<think>\textbackslash n} seeded immediately after the assistant role marker, matching their training distribution. Instruction-tuned baselines ({\sc Qwen3-0.6B}, {\sc LFM2-350M}, {\sc Gemma-3-270M-IT}, {\sc SmolLM2-360M}) instead receive a short instruction system prompt : \textit{``Answer the multiple choice question by reasoning step by step, then give your final answer as a single letter (A, B, C, or D).''} for MCQ, and \textit{``Answer the question concisely and accurately.''} for open-ended. {\sc Qwen3}, itself a thinking model, additionally has \texttt{<think>\textbackslash n} seeded for both tasks.

For thinking models, the \texttt{<think>...</think>} trace is stripped before scoring. Generation uses temperature 0.1. MCQ benchmarks with more than 1000 questions are subsampled to 1000, picked with seed 42. MCQ answers are extracted by a multi-level regex cascade (letter at the start of the answer, ``Answer: X'' patterns, line-leading letter, choice-text match, and a last-resort standalone capital) and scored as letter-vs-gold. Open-ended answers are graded with a LLM-as-judge approach using {\sc Qwen3-30B-A3B}, classifying each response as \texttt{CORRECT}, \texttt{INCORRECT}, \texttt{NOT\_ATTEMPTED}, or \texttt{UNPARSEABLE}; reported accuracy is the fraction of \texttt{CORRECT} verdicts.

\section{Tool calling results}\label{app:tool-calling}

\begin{table}[h]
\centering
\caption{\textbf{Tool calling on BFCL v2.} Micro: accuracy over all items; macro: mean over categories. The base model emits no tool calls and scores only on the irrelevance categories.}
\label{tab:tool-calling}
\small
\begin{tabular}{lrrr}
\toprule
\textbf{Model} & \textbf{Training tokens} & \textbf{Micro} & \textbf{Macro} \\
\midrule
\textsc{Baguettotron}-600M + tool calling & 158B + 1.5B & \textbf{53.1} & \textbf{53.1} \\
\textsc{Baguettotron}-350M + tool calling & 200B + 1.5B & 47.7 & 46.9 \\
FunctionGemma-270M~\citep{functiongemma2025} & 6T & 49.2 & 45.3 \\
\textsc{Baguettotron}-600M (base) & 158B & 28.0 & 11.1 \\
\bottomrule
\end{tabular}
\end{table}

\clearpage
\section*{NeurIPS Paper Checklist}

\begin{enumerate}

\item {\bf Claims}
    \item[] Question: Do the main claims made in the abstract and introduction accurately reflect the paper's contributions and scope?
    \item[] Answer: \answerYes{}
    \item[] Justification: Each of our contributions has a reference to the section/subsection discussing the claims, and the same claims are made in the abstract.

\item {\bf Limitations}
    \item[] Question: Does the paper discuss the limitations of the work performed by the authors?
    \item[] Answer: \answerYes{}
    \item[] Justification: See section Limitations and future work

\item {\bf Theory assumptions and proofs}
    \item[] Question: For each theoretical result, does the paper provide the full set of assumptions and a complete (and correct) proof?
    \item[] Answer: \answerNA{}
    \item[] Justification: No theoretical proofs, all empirical results.

    \item {\bf Experimental result reproducibility}
    \item[] Question: Does the paper fully disclose all the information needed to reproduce the main experimental results of the paper to the extent that it affects the main claims and/or conclusions of the paper (regardless of whether the code and data are provided or not)?
    \item[] Answer: \answerYes{}
    \item[] Justification: While the manuscript should be enough to reproduce our results, we released models and the Synth dataset to aid reproductions.

\item {\bf Open access to data and code}
    \item[] Question: Does the paper provide open access to the data and code, with sufficient instructions to faithfully reproduce the main experimental results, as described in supplemental material?
    \item[] Answer: \answerYes{}
    \item[] Justification: The supplementary material includes a 5{,}000-row sample of \synth{} and the FActScore evaluation code. Anonymized \textsc{Baguettotron}-350M weights are hosted at \url{https://anonymous-hf.up.railway.app/a/3e0gcxurz29r/}. The full \synth{} corpus (CC BY 4.0) and the remaining \textsc{Baguettotron} models (Apache-2.0) will be released with the camera-ready version. Training and data-generation code is not released; the dataset construction (\autoref{ch:synth}, \autoref{app:query-constraints}), training configurations (\autoref{app:training-runs}), and evaluation protocol (\autoref{app:factscore}, \autoref{app:benchmarks}) are documented in the paper at a level sufficient for re-implementation.

\item {\bf Experimental setting/details}
    \item[] Question: Does the paper specify all the training and test details (e.g., data splits, hyperparameters, how they were chosen, type of optimizer) necessary to understand the results?
    \item[] Answer: \answerYes{}
    \item[] Justification: Pre-training hyperparameters (optimizer, schedule, batch size, sequence length, steps, token budget) are reported per model in \autoref{app:training-runs}; evaluation prompts, sampling parameters, and scoring are in \autoref{app:benchmarks} and \autoref{app:factscore}.

\item {\bf Experiment statistical significance}
    \item[] Question: Does the paper report error bars suitably and correctly defined or other appropriate information about the statistical significance of the experiments?
    \item[] Answer: \answerYes{}
    \item[] Justification: FActScore comparisons (\autoref{ch:memorization}) report cluster-bootstrap confidence intervals over 10{,}000 resamples and paired Wilcoxon signed-rank tests on per-entity precision (\autoref{app:factscore}). Pre-training and benchmark accuracies are reported as single runs; repeating full pre-training across seeds is computationally infeasible at this scale, and we instead report token-budget trajectories and per-benchmark variance across 26 MCQ tasks (\autoref{app:benchmarks}).

\item {\bf Experiments compute resources}
    \item[] Question: For each experiment, does the paper provide sufficient information on the computer resources (type of compute workers, memory, time of execution) needed to reproduce the experiments?
    \item[] Answer: \answerYes{}
    \item[] Justification: Hardware (16$\times$H100 GPUs per run), wall-clock time, MFU, and token budgets are reported per model in \autoref{app:training-runs}. Generator fine-tuning was performed on a single H100. The architecture-ablation sweep used the same 16$\times$H100 setup over 10B tokens per configuration.
    
\item {\bf Code of ethics}
    \item[] Question: Does the research conducted in the paper conform, in every respect, with the NeurIPS Code of Ethics \url{https://neurips.cc/public/EthicsGuidelines}?
    \item[] Answer: \answerYes{}
    \item[] Justification: The research conforms with the NeurIPS Code of Ethics. No human subjects are involved, all source data (Wikipedia, Wikibooks, Common Corpus) is used under permissive licenses, and released artifacts are shared under permissive licenses (\synth{}: CC BY 4.0; \textsc{Baguettotron}: Apache-2.0).

\item {\bf Broader impacts}
    \item[] Question: Does the paper discuss both potential positive societal impacts and negative societal impacts of the work performed?
    \item[] Answer: \answerYes{}
    \item[] Justification: Positive impacts are discussed throughout: substantially lower training-token and compute requirements (\autoref{ch:baguettotron}) lower the barrier to training open small models, and the released \synth{} corpus enables a fully open and reproducible pre-training pipeline that does not depend on web-scraped data of uncertain provenance. We do not discuss negative societal impacts in detail because the released models are small ($\leq 1.05$B active parameters) with limited generative capabilities relative to frontier systems, and the dataset is grounded in publicly licensed encyclopedic sources.
    
\item {\bf Safeguards}
    \item[] Question: Does the paper describe safeguards that have been put in place for responsible release of data or models that have a high risk for misuse (e.g., pre-trained language models, image generators, or scraped datasets)?
    \item[] Answer: \answerNA{}
    \item[] Justification: The released models are small ($\leq 1.05$B active parameters) and the training corpus is grounded in publicly licensed encyclopedic sources rather than web scrape, so the released artifacts pose limited dual-use risk. \synth{}'s third-party integrity/safety scores (\autoref{app:propella}) place it at the top of the open-corpus distribution alongside Wikipedia-derived data.

\item {\bf Licenses for existing assets}
    \item[] Question: Are the creators or original owners of assets (e.g., code, data, models), used in the paper, properly credited and are the license and terms of use explicitly mentioned and properly respected?
    \item[] Answer: \answerYes{}
    \item[] Justification: All third-party datasets, models, and tools used in the paper are cited at first use. Seed sources are Wikipedia and Wikibooks (CC BY-SA 4.0) and Common Corpus~\citep{langlais2026common} for tokenizers; baseline corpora referenced in the data-quality comparison (Nemotron-CC, FinePDFs, FineWeb-2, FineWiki, HPLT-4) are cited with their original references. Generator base models include Gemma-3-PT (Gemma Terms of Use) and Qwen3 (Apache-2.0); the \texttt{bge-m3} retriever (MIT) and \texttt{torchtitan}/\texttt{Nanotron} training stacks are also cited. All assets are used in compliance with their respective licenses.

\item {\bf New assets}
    \item[] Question: Are new assets introduced in the paper well documented and is the documentation provided alongside the assets?
    \item[] Answer: \answerYes{}
    \item[] Justification: We release the \synth{} dataset under CC BY 4.0 and the \textsc{Baguettotron} model suite (\textsc{Monad}-56M, \textsc{Baguettotron}-350M/600M, and \textsc{Baguettotron}-MoE) under Apache-2.0. For anonymous review, the supplementary zip contains a 5{,}000-row \synth{} sample and the FActScore evaluation code, and \textsc{Baguettotron}-350M is hosted anonymously at \url{https://anonymous-hf.up.railway.app/a/3e0gcxurz29r/}. Dataset construction, languages, task mix, and constraint priors are documented in \autoref{ch:synth} and \autoref{app:query-constraints}; per-model architectures, tokenizers, and training recipes in \autoref{ch:baguettotron} and \autoref{app:training-runs}. Each released artifact will ship with a model/dataset card at the camera-ready stage.

\item {\bf Crowdsourcing and research with human subjects}
    \item[] Question: For crowdsourcing experiments and research with human subjects, does the paper include the full text of instructions given to participants and screenshots, if applicable, as well as details about compensation (if any)?
    \item[] Answer: \answerNA{}
    \item[] Justification: The paper does not involve crowdsourcing or research with human subjects.

\item {\bf Institutional review board (IRB) approvals or equivalent for research with human subjects}
    \item[] Question: Does the paper describe potential risks incurred by study participants, whether such risks were disclosed to the subjects, and whether Institutional Review Board (IRB) approvals (or an equivalent approval/review based on the requirements of your country or institution) were obtained?
    \item[] Answer: \answerNA{}
    \item[] Justification: The paper does not involve crowdsourcing or research with human subjects.

\item {\bf Declaration of LLM usage}
    \item[] Question: Does the paper describe the usage of LLMs if it is an important, original, or non-standard component of the core methods in this research? Note that if the LLM is used only for writing, editing, or formatting purposes and does \emph{not} impact the core methodology, scientific rigor, or originality of the research, declaration is not required.
    \item[] Answer: \answerYes{}
    \item[] Justification: Used for LLM-as-a-Judge in \autoref{ch:memorization}, used for synthetic data generation in \autoref{ch:synth}

\end{enumerate}

\end{document}